\pdfoutput=1
\documentclass[11pt]{article}

\usepackage[final]{acl}

\usepackage{times}
\usepackage{latexsym}
\usepackage{enumitem}

\usepackage[T1]{fontenc}
\usepackage[utf8]{inputenc}

\usepackage{microtype}

\usepackage{inconsolata}

\usepackage{graphicx}
\usepackage{amsmath}
\usepackage{amssymb}
\usepackage{booktabs}
\usepackage{bm}
\usepackage{amsfonts}
\usepackage{multirow}
\usepackage{subcaption}
\usepackage[table]{xcolor}
\usepackage{comment}
\usepackage{mathtools}
\usepackage{dsfont}
\definecolor{medcolor}{RGB}{255, 214, 171} %medical llms
\definecolor{reasoncolor}{RGB}{192, 231, 237} %reasoning
\usepackage{fvextra}
\usepackage{fancyvrb}
\usepackage{tabularx} 

\newcommand\benchmark{DeVisE} % Benchmark dataset name suggestions:
\definecolor{niceorange}{HTML}{F86624}
\definecolor{niceblue}{HTML}{072AC8}

\newcommand{\parahead}[1]{\vspace{0.5mm}\noindent\textbf{#1}.\ }

\DeclareMathOperator*{\EX}{\mathbb{E}}% expected value
\newcommand\scalemath[2]{\scalebox{#1}{\mbox{\ensuremath{\displaystyle #2}}}}
\DeclareMathOperator{\KL}{KL}
\DeclarePairedDelimiterX{\infdivx}[2]{(}{)}{%
  #1\;\delimsize\|\;#2%
}
\newcommand{\infdiv}{\KL\infdivx}

\newcommand\varage[0]{\texttt{age}}
\newcommand\vargender[0]{\texttt{gender}}
\newcommand\varethnicity[0]{\texttt{ethnicity}}
\newcommand\varheartrate[0]{\texttt{heart rate}}
\newcommand\varsystolicbloodpressure[0]{\texttt{systolic blood pressure}}
\newcommand\vardiastolicbloodpressure[0]{\texttt{diastolic blood pressure}}
\newcommand\varbodytemperature[0]{\texttt{body temperature}}
\newcommand\varrespiratoryrate[0]{\texttt{respiratory rate}}
\newcommand\varoxygensaturation[0]{\texttt{oxygen saturation}}
\newcommand\varheartrateacronym[0]{\texttt{HR}}
\newcommand\varsystolicbloodpressureacronym[0]{\texttt{SBP}}
\newcommand\vardiastolicbloodpressureacronym[0]{\texttt{DBP}}
\newcommand\varbodytemperatureacronym[0]{\texttt{T}}
\newcommand\varrespiratoryrateacronym[0]{\texttt{RR}}
\newcommand\varoxygensaturationacronym[0]{\texttt{O2Sat}}
\DeclareMathOperator{\xcounterfactualspervariable}{\mathbf{\tilde{x}}_{i}^{(v \leftarrow a)}}

\usepackage{xspace}
\newcommand{\phiF}{Phi4-14B\xspace}
\newcommand{\meditron}{Meditron3-Phi4-14B\xspace}
\newcommand{\llama}{LLaMA-3.3-Instruct-70B\xspace}
\newcommand{\obllm}{OpenBioLLM-70B\xspace}
\newcommand{\deepseek}{DeepSeek-R1-Distill-70B\xspace}
\newcommand{\qwen}{Qwen-2.5-Instruct-72B\xspace}
\newcommand{\gptoss}{GPT-OSS-120B\xspace}
\newcommand{\gptmini}{GPT-4.1-mini\xspace}

\title{\benchmark{}: Towards the Behavioral Testing of \\ Medical Large Language Models}

\author{
  \textbf{Camila Zurdo Tagliabue\textsuperscript{1}}, \quad
  \textbf{Heloisa Oss Boll\textsuperscript{1,2}}, \\
  \textbf{Aykut Erdem\textsuperscript{3}}, \quad
  \textbf{Erkut Erdem\textsuperscript{4}}, \quad
  \textbf{Iacer Calixto\textsuperscript{1,2}} \\
  \\
  \textsuperscript{1}Department of Medical Informatics, Amsterdam UMC, University of Amsterdam, \\ Amsterdam, The Netherlands \\
  \textsuperscript{2}Amsterdam Public Health, Methodology, Amsterdam, The Netherlands\\
  \textsuperscript{3}Koç University, Istanbul, Turkey \qquad
  \textsuperscript{4}Hacettepe University, Ankara, Turkey \\
  \\
}

\begin{document}
\maketitle
\begin{abstract}
Large language models (LLMs) are increasingly applied in clinical decision support, yet current evaluations rarely reveal whether their outputs reflect genuine medical reasoning or superficial correlations. We introduce \benchmark{} (\textbf{\underline{De}}mographics and \textbf{\underline{Vi}}tal \textbf{\underline{s}}igns \textbf{\underline{E}}valuation), a behavioral testing framework that probes fine-grained clinical understanding through controlled counterfactuals. Using intensive care unit (ICU) discharge notes from MIMIC-IV, we construct both raw (real-world) and template-based (synthetic) variants with single-variable perturbations in demographic (age, gender, ethnicity) and vital sign attributes. We evaluate eight LLMs, spanning general-purpose and medical variants, under zero-shot setting. Model behavior is analyzed through (1) input-level sensitivity, capturing how counterfactuals alter perplexity, and (2) downstream reasoning, measuring their effect on predicted ICU length-of-stay and mortality. Overall, our results show that standard task metrics obscure clinically relevant differences in model behavior, with models differing substantially in how consistently and proportionally they adjust predictions to counterfactual perturbations.\footnote{Our benchmark is available at \url{https://github.com/camztag/DeVisE}}
%Large language models (LLMs) are increasingly applied in clinical decision support, yet current evaluations rarely reveal whether their outputs reflect genuine medical reasoning or superficial correlations. We introduce \benchmark{} (\textbf{\underline{De}}mographics and \textbf{\underline{Vi}}tal \textbf{\underline{s}}igns \textbf{\underline{E}}valuation), a behavioral testing framework that probe fine-grained clinical understanding through controlled counterfactuals. Using intensive care unit (ICU) discharge notes from MIMIC-IV, we construct both raw (real-world) and template-based (synthetic) variants with single-variable perturbations in demographic (age, gender, ethnicity) and vital sign attributes. We evaluate five LLMs, spanning general-purpose and medical variants, under zero-shot and fine-tuned settings. Model behavior is analyzed through (1) input-level sensitivity, capturing how counterfactuals alter likelihoods, and (2) downstream reasoning, measuring their effect on predicted ICU length-of-stay. Zero-shot models exhibit more coherent and clinically aligned reasoning patterns, whereas fine-tuned ones are stabler but less sensitive to meaningful changes. Demographic factors subtly yet consistently shape predictions, emphasizing the need for fairness-aware evaluation in medical LLMs. 
%This work highlights the utility of behavioral testing in exposing the reasoning strategies of clinical LLMs and informing the design of safer, more transparent medical AI systems.
%commented out for anonymity
\end{abstract}

\section{Introduction}
\label{sec:intro}
Large language models (LLMs) are increasingly applied in the medical domain and show strong performance across clinical tasks~\cite{ gu2021domain, mcduff2023towards, van2024adapted, singhal2025toward}. However, conventional medical benchmarks~\cite{xu2023medgpteval, bakhshandeh2023benchmarking, yao2024medqa} offer limited insight into how clinically grounded is an LLM's manipulation of key clinical variables when making predictions, such as 
%answering a medical question or
predicting the risk of mortality of a patient in the hospital~\cite{vanAken2021, MacPhail2024, Jullien2024, van2024adapted, singhal2025toward}.
A key question is whether 
%In other words, LLMs are black-boxes and we do not know whether an 
an LLM makes use of clinical concepts similarly to how a human clinician would, or whether it relies on biases, shortcuts, and spurious correlations. %Models that excel on these metrics can still fail in settings requiring fine-grained clinical reasoning~\cite{Aguiar2024, CeballosArroyo2024}.

\begin{figure*}[t!]
    \centering
    \includegraphics[width=0.88\linewidth]{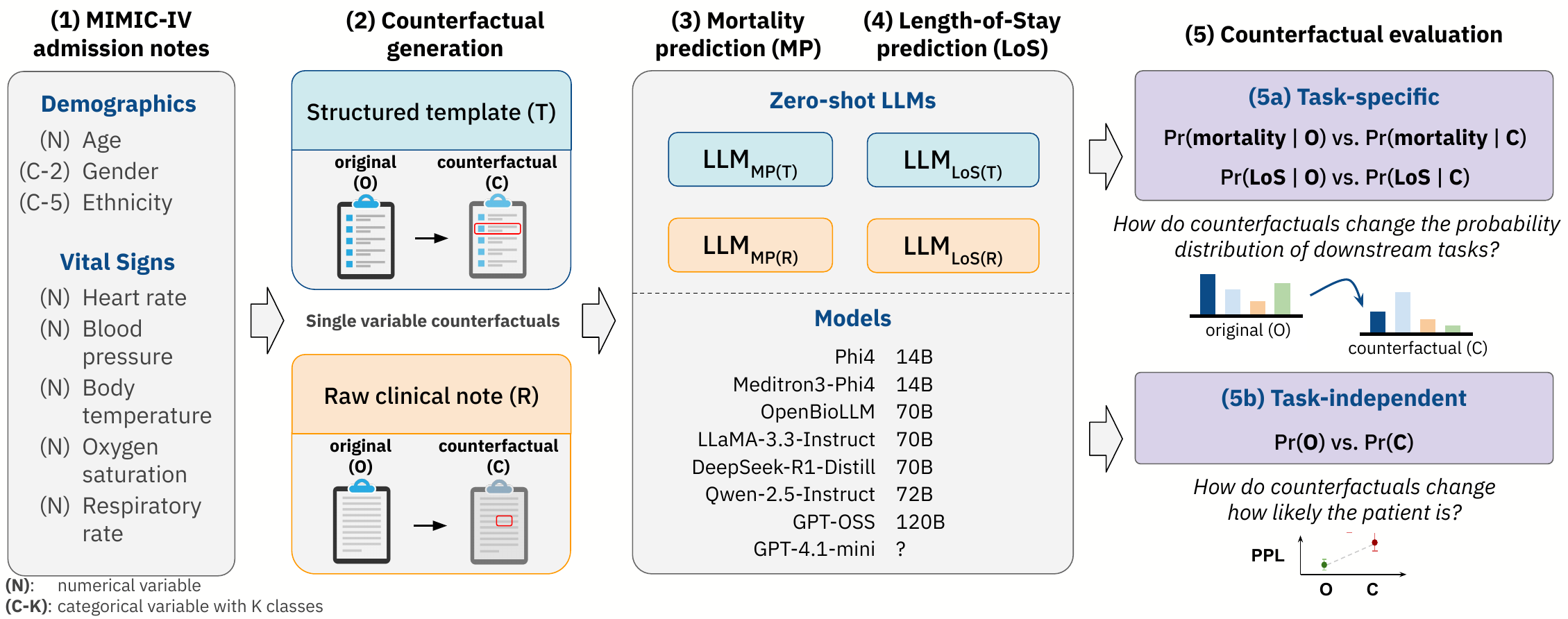}
    \caption{Overview of \textbf{\benchmark} in 5 steps: (1) We create admission notes using MIMIC-IV discharge summaries; (2) We create counterfactuals for template-based and raw notes; (3--4) We use original and counterfactual notes to predict mortality and LoS in a zero-shot setting, and (5) We evaluate and compare LLMs in (5a) a task-specific setting---on how counterfactuals affect mortality and LoS probabilities---and in (5b) a task-independent setting, by comparing how LLM perplexity change between original and counterfactual examples.}

%\caption{Overview of \textbf{\benchmark} in 5 steps: (1) we create admission notes using MIMIC-IV discharge summaries, (2) create counterfactuals for template-based and raw notes, (3--4) use original notes to adapt (train) zero-shot (fine-tuned) LLMs to predict mortality and LoS, and (5) evaluate zero-shot and fine-tuned LLMs on mortality and LoS prediction and in a task-independent setting, e.g., comparing LLMs' perplexity for original vs. counterfactuals.}

   %\caption{\textbf{Overview of \benchmark}. We construct a dataset of 1,000 MIMIC-IV discharge summaries with manually validated single-variable counterfactuals targeting key clinical attributes: \textbf{demographics} ({age}, {gender}, {ethnicity}) and \textbf{vital signs} ({heart rate}, {respiration rate}, {oxygen saturation}, {temperature}, {blood pressure}). We probe LLMs' sensitivity to input perturbations (e.g., ``How does a change in age affect the model's belief in this note?'') and downstream effects (e.g.,  ``How does this change affect predicted length-of-stay?'').}
    \label{fig:f1}
\end{figure*}
%To address this gap, behavioral testing offers a complementary evaluation paradigm. Originating in software engineering, it involves assessing a system’s behavior through its input–output responses, without requiring access to its internal workings~\cite{beizer1996black}. In Natural Language Processing (NLP), \citet{ribeiro2020beyond} adapted this approach to evaluate linguistic capabilities across a range of tasks.
%Behavioral testing (BT) is an evaluation strategy that assesses a system’s behavior without white-box access to a model simply by checking the model's input--output responses~\cite{beizer1996black}.

%To illustrate how we address this issue, let us use the example of a patient (referred to as the \textit{`original patient'}) in which a clinically meaningful variable such as the \varheartrate{} is changed from 89 / `normal' to 120 / `high' (referred to as the \textit{`counterfactual patient'}).
To illustrate how we address this issue, let us use the example of an \textit{`original patient'} in which a clinically meaningful variable such as the \varheartrate{} is changed from 89 bpm (`normal') to 120 bpm (`high'), yielding a \textit{`counterfactual patient'}.
We build on the intuition that, in such cases, an LLM should both adjust 1) \textit{the probability of the patient clinical profile}, such that the counterfactual patient profile should become \textit{less likely the further it gets from the original}, and 2) \textit{the probability of associated outcomes}, such as predicting a \textit{higher risk of mortality} for the counterfactual compared to the original patient, since the clinical condition of the patient deteriorated when moving from  \varheartrate{} 89 bpm (`normal') to 120 bpm (`high').
%Had the input variable instead changed in the opposite direction---i.e., causing the patient clinical condition to improve---, we would have expected the same LLM to adjust and predict a \textit{lower risk of mortality}.
%In other words, we expect a \textit{monotonic} and \textit{smooth} relationship between input clinical variables and outcome.

%We build on the idea (Fig.~\ref{fig:f1}) that when a clinically meaningful variable (e.g., blood pressure) changes from normal to very high, an LLM should adjust the likelihood of associated outcomes, such as a longer predicted hospital stay. Altering a single attribute may also yield an implausible patient profile, and the model’s predictions should reflect that inconsistency.

%Example questions include ``How does a change in the patient age affect the model?''---and downstream effects---e.g., ``How does this change affect predicted length-of-stay?''.

In this work, we propose \textbf{\benchmark}, a \textit{clinically-grounded evaluation framework for medical LLMs based on behavioural testing and counterfactual evaluation}.
%\textbf{\benchmark} is a behavioral testing benchmark for clinical NLP based on 
%It uses MIMIC-IV discharge summaries \cite{johnson2023mimic} and focuses on \textit{minimally differing counterfactuals for core clinical variables}. %: \textit{demographics} (age, gender, ethnicity) and \textit{vital signs} (heart rate, body temperature, respiration rate, oxygen saturation, and blood pressure).
We show an overview of \benchmark{} in Figure~\ref{fig:f1}.
(1) \textbf{Admission notes:} We create admission notes for intensive care unit (ICU) patients from MIMIC-IV~\cite{johnson2023mimic}
discharge summaries following~\citet{van2021clinical,rohr2024revisiting}.
We extract key clinical variables: demographics (\varage{}, \vargender{}, \varethnicity{}) from MIMIC's structured data, and vital signs (\varheartrate{}, \varsystolicbloodpressure{}, \vardiastolicbloodpressure{}, \varbodytemperature{}, \varoxygensaturation{}, \varrespiratoryrate{}) from the admission notes.
We choose these variables due to their relevance to the two downstream tasks we use in our evaluation ~\citep{hempel2023prediction, candel2022association} as well as for their low missingness rate.
We use both raw admission notes and a template-based version of the same notes whereby only the variables of interest are included.
(2) \textbf{Counterfactual generation:} For each patient, we create single-variable counterfactuals for all variables, e.g., change a patient’s \varheartrate{} from 89 to 120, while keeping everything else unchanged.
(3–4) \textbf{Mortality} and \textbf{Length-of-stay prediction:} We apply LLMs in a zero-shot setting to predict two downstream tasks: mortality
and length-of-stay (LoS) in the ICU.
%(3–4) \textbf{Mortality} and \textbf{Length-of-stay prediction:} We apply LLMs in a zero-shot and fine-tuned setting to predict two downstream tasks: mortality
%and length-of-stay (LoS) in the ICU.
(5) \textbf{Counterfactual evaluation:}
In (5a) \textbf{Task-specific evaluation}, we investigate how counterfactuals change the probability of the two downstream tasks. For counterfactuals implying a change in the patient condition, we expect the LLM to reflect that in the downstream probabilities.
For instance, in our example the patient condition deteriorates with the \varheartrate{} changing from  89 / `normal' to 120 / `high'.
We thus expect the LLM to predict a higher risk of mortality and longer LoS in the ICU for the counterfactual relative to the original patient. 
In (5b) \textbf{Task-independent evaluation}, we quantify to what extent counterfactuals affect how likely patients are, e.g., we compare the perplexity of the original vs. counterfactual patient notes. We expect that the further a counterfactual variable is from the true value of the clinical variable, the lower the probability the LLM should assign to the counterfactual.

%We create raw admission notes from MIMIC-IV discharge summaries and a template-based version of these notes whereby only the variables of interest are included.
%\benchmark{} consists of $1,000$ raw clinical notes, their template-based version, as well as counterfactuals for both raw and template-based notes (all manually validated).
%By assessing model behavior without white-box access to a model~\cite{beizer1996black}, behavioral testing (BT) enables transparent evaluation of whether predictions align with clinically meaningful reasoning, a key requirement since patient outcomes can depend on subtle variations in only a few clinical variables~\cite{downey2017strengths, alghatani2021predicting, herasevich2022impact}.
%Moreover, by analyzing both raw and template-based notes we disentangle the effect of linguistic complexity from variable sensitivity~\cite{adnan2020role, zhang2020combining}. To our knowledge, prior work has mostly focused on structures templates or fully synthetic inputs~\cite{vanAken2021, MacPhail2024, Aguiar2024, rajagopal2021template}, leaving the behavioral evaluation of medical LLMs on realistic text largely unexplored.

Our main contributions are:
\begin{itemize}[noitemsep,topsep=0pt]
    \item We introduce \textbf{\benchmark} (\textbf{\underline{De}}mographics and \textbf{\underline{Vi}}tal \textbf{\underline{s}}igns \textbf{\underline{E}}valuation), a benchmark for clinical NLP that applies behavioral testing with \textit{minimally differing counterfactuals} across demographic and vital sign variables.
    \item We release a dataset of $1,001$ raw and $1,001$ template-based validated notes, each with corresponding counterfactuals, enabling controlled evaluation of model robustness to clinically meaningful counterfactuals. %\footnote{Following MIMIC data usage agreement, in \url{https://physionet.org}}
    \item We compare eight state-of-the-art LLMs across dimensions: template-based vs. raw clinical note,  task-independent vs. task-specific performance, medical vs. general-purpose vs. reasoning-oriented LLMs.
%\item We compare eight state-of-the-art LLMs across multiple dimensions: fine-tuned vs. zero-shot, medical vs. general-purpose, reasoning-oriented vs. non-reasoning, and task-independent vs. task-specific performance (Fig.~\ref{fig:f1}).
\end{itemize}

%We aim to provide deeper insights into model performance and contribute to a more reliable assessment of medical LLMs, helping clinicians and researchers understand their strengths, limitations, and potential areas of improvement.
%This work takes an initial step toward behavioral testing of clinical LLMs. 
While not exhaustive, \benchmark{} provides a controlled and clinically meaningful testbed for probing reasoning patterns, paving the way for broader evaluations and safer deployment in real-world medical settings.
\section{Method}
\benchmark{} is designed as a test-only clinical evaluation benchmark and includes three main tests, illustrated in Figure~\ref{fig:f1}: two task-specific tests measuring how \textit{mortality} and \textit{LoS prediction} changes under counterfactuals, and one task-independent test measuring how \textit{patient clinical profile probability} changes under counterfactuals.
LLMs are not trained on original or counterfactual examples we generate in any way, and in that sense \textit{\benchmark{} is a zero-shot benchmark}.

\paragraph{Background}
Let $\mathcal{P} \in \mathcal{D}$ denote the set of patients in a dataset $\mathcal{D}$ (detailed further in \S\ref{sec:source_dataset}) and $\mathcal{V}$ the set of clinical variables (explained in \S\ref{sec:clinical_variables}).
Each variable $v \in \mathcal{V}$ takes values in a domain $\mathcal{X}_v$ defined by ranges (numerical variables) or classes (categorical variables) defined in Tables~\ref{tab:demo-var}--\ref{tab:vitals-var}.
%We introduce the dataset we use in \S~\ref{sec:source_dataset} and the clinical variables in \S~\ref{sec:clinical_variables}.
%In \S~\ref{sec:data_preparation} we explain how we build template-based notes from raw free-text notes as well as how we map clinical variables onto \textit{clinically-grounded severity scales} used in our analysis.
In \S\ref{sec:data_preparation} we explain how we map numerical variables $v \in \mathcal{V}$ onto \textit{clinically-grounded severity scales} for our analysis.
Patient $i \in \mathcal{P}$ is represented by a feature vector $\mathbf{x}_i = (x_{iv})_{v \in \mathcal{V}} \in \mathcal{X} = \prod_{v \in \mathcal{V}} \mathcal{X}_v$.

For each patient $i$ and variable $v$, let $\mathcal{A}_{iv} \subset \mathcal{X}_v$ denote the set of \textit{alternative values} or \textit{possible counterfactuals}, with $a \neq x_{iv}$ for all $a \in \mathcal{A}_{iv}$. A \textit{single-variable counterfactual} for patient $i$ 
%corresponding to changing variable $v$ to value $a \in \mathcal{A}_{iv}$ 
is the feature vector obtained by replacing the original value of $v$ in $\mathbf{x}_i$ with $a$, while keeping all other variables unchanged; we denote this vector by $\xcounterfactualspervariable$. 
The set of all counterfactuals for patient $i$ and variable $v$ is
$\mathcal{C}_{iv} = \{ \xcounterfactualspervariable \mid v \in \mathcal{V}, \forall a \in \mathcal{A}_{iv} \}$,
and the set of all counterfactuals for all variables for a patient $i$ is
$\mathcal{C}_i = \{ \xcounterfactualspervariable \mid \forall v \in \mathcal{V}, \forall a \in \mathcal{A}_{iv} \}$.
%and the set of all counterfactuals for patient $i$ is $\mathcal{C}_i =  \bigcup_{v \in \mathcal{V}} \mathcal{C}_{iv} $.
We explain how we generate counterfactuals in \S~\ref{sec:counterfactuals} and detail our manual validation of the original and counterfactual patients' data in \S~\ref{sec:data_validation}.

Each patient $i$ is associated with two categorical outcomes: mortality $y_i^{\text{mort}} \in \mathcal{Y}^{\text{mort}}$ and length-of-stay $y_i^{\text{los}} \in \mathcal{Y}^{\text{los}}$,
%with $\mathcal{Y}^\text{mort} \in \{0,1\}$ and $\mathcal{Y}^\text{los} \in \{0,1,2,3\}$ 
discussed further in \S~\ref{sec:mortality_prediction} and \S~\ref{sec:los_prediction}, respectively.
Let $f_\theta : \mathcal{X} \to \{\mathcal{Y}^{\text{mort}}, \mathcal{Y}^{\text{los}}\}$ denote a model with parameters $\theta$; we write $f_\theta(\mathbf{x}_i)$ and $f_\theta(\xcounterfactualspervariable)$ for predictions on observed and counterfactual instances, respectively.
%, and define the counterfactual effect as $\Delta_i^{(v \leftarrow a)} = f(\mathbf{x}_i^{(v \leftarrow a)}) - f(\mathbf{x}_i)$.

%We first introduce the dataset used to source examples for \benchmark{} (\S~\ref{sec:source_dataset}).
%We then explain the choice of clinical variables we use in our counterfactual evaluation (\S~\ref{sec:clinical_variables}), how we build template-based notes from raw free-text notes as well as how we map clinical variables onto \textit{clinically-grounded severity scales} used in our analysis (\S~\ref{sec:data_preparation}).
%Moreover, we explain how we generate counterfactuals for all patients in our test set (\S~\ref{sec:counterfactuals}) and provide details about our manual validation of the original and counterfactual patients' data (\S~\ref{sec:data_validation}).
%Finally, we introduce the mathematical notation we use throughout the paper (\S~\ref{sec:notation}).

\subsection{Source dataset}
\label{sec:source_dataset}
We source our data from the MIMIC-IV database~\cite{johnson2023mimic,Johnson2023-gt}, which includes $65,366$ intensive care unit (ICU) patients from the Beth Israel Deaconess Medical Center in Boston, MA, USA, and documents the full course of a hospital stay.
MIMIC-IV includes patients' structured data (such as demographics, prescribed medication, vital signs, and lab results) and also free-text discharge notes.
Discharge notes are semi-structured and include information about the patient medical condition prior to admission to the ICU, as well as about the patient's ICU stay.
%\red {splits even if no fine tuning?}
%We followed a split \%85-15 (train-test) and splitting the 85 into 1/3 for validation. Our dataset comprises 35,944 training, 6,584 validation, and 7,730 test set samples. 
Our benchmark includes 1,001 randomly selected admission notes from MIMIC-IV's test set, representative of the adult population in MIMIC-IV: 45\% female, mean age 64$\pm$17 years, and 31\% non-white (see Table~\ref{tab:population_stats}). 
%\red{Mention in one sentence what is the split we use: from previous work? Then cite. We should be sourcing all examples in \benchmark{} from MIMIC-IV test set. The validation set is only relevant for experiments with few-shot learning, since we will source few-shot demonstrations from the validation set. We never use the training set in our experiments.}

\paragraph{From discharge notes to admission notes}
Since in our task-specific tests we predict mortality and LoS in the ICU, we preprocess discharge summaries to remove all information pertinent to the current ICU stay.
We follow \citet{van2021clinical,rohr2024revisiting} and keep only information available at the time of admission in what we refer to as \textbf{admission notes}.
Specifically, we preprocess and clean discharge summaries retaining only the following fields (and redacting everything else): \textit{chief complaint}, \textit{present illness}, \textit{medical history}, \textit{admission medications}, \textit{allergies}, \textit{physical exam}, \textit{family history}, and \textit{social history}.

\paragraph{Raw admission notes vs. template-based notes}
Raw admission notes may include a wide range of information not directly relevant for particular downstream tasks.
For that reason, we create a template-based version of a patient's admission note whereby only the variables we are interested in and their value are included (see \S~\ref{sec:demo-notes-raw} and \S~\ref{sec:demo-notes-templatebased} for examples of a raw note and its corresponding template-based version).
Template-based notes allow us to isolate model responses to variable changes from confounding arising from natural variation and noise in the original raw text notes.

When using template-based notes, we shuffle the variables' serialisation order before feeding a note to an LLM.
%both during training and inference.
We discuss experiments with raw- and template-based notes in detail later in \S~\ref{sec:experimental_setup}.
%\red{Explain why we create template-based notes: what are our expectations in terms of findings with template-based vs. raw notes?}
%\red{Explain how we create template-based notes from raw notes.}
%\red{Explain the shuffling of the variable order in the template-based notes.}

%\paragraph{Splits}
%\red{Explain what are the splits we use (cite other work if we use splits proposed in previous work). Explain if the splits are made at the patient level.}

%\paragraph{Inclusion criteria}
%We apply the cohort selection criteria below for all splits.
%First of all, we only use the first ICU stay for the patients in our cohort.
%For patients with multiple ICU stays, when these occur less than 48 hours apart we merge them into a single ICU stay; if two consecutive stays are separated by more than 48 hours, we treat them as distinct ICU admissions.
%
%We predict mortality (binary) and length-of-stay (categorical) 24 hours after admission at the ICU, and therefore remove any patients who die within the first 24 hours or who have an ICU stay shorter than 24 hours.

%\paragraph{Descriptive statistics of the structured and unstructured data}

\begin{table}[t!]
\centering
\resizebox{.48\textwidth}{!}{
\begin{tabular}{lll}
\toprule
\textbf{Variable} (\textbf{Range} $\mathcal{X}_v$) & \textbf{Range per class} & \textbf{Classes}\\
%\textbf{(\% Missing)} \\
\midrule
\multirow{4}{*}{\parbox{1.4cm}{Age\\(18--91)}} & 18--35 & young adults \\
             & 36--55 & middle aged adults \\
             & 56--75 & older adults \\
             & 76--91 & elderly \\
\midrule
\multirow{2}{*}{\parbox{1.5cm}{Gender (N/A)}}          &--- & female \\
                &---& male \\
\midrule
\multirow{5}{*}{\parbox{2cm}{Ethnicity (N/A)}}          &---& Asian \& Pacific \\
              &---& Black \\
              &---& Hispanic/Latino \\
              &---& Other/Unknown \\
              &---& White \\
\bottomrule
\end{tabular}}
\caption{Demographic variables used in \benchmark{}.
%along with their missingness. 
%Age ranges are grouped into life-stage categories to improve interpretability and support counterfactual evaluation.
%\red{We group age ranges ... [Explain how we group ranges: based on a clinical guideline? Previous work?]}
%$^\dagger$: 
%Age is a numerical variable and classes are not included in MIMIC-IV. 
%We explain the mapping of age from numerical to categorical in \S \ref{sec:data_preparation}. 
}
\label{tab:demo-var}
\end{table}

\subsection{Clinical variables}
\label{sec:clinical_variables}
%\begin{itemize}
%    \item Demographics \& Vital signs
%    \item Missingness
%\end{itemize}

\paragraph{Demographics}
We use \varage{}, \vargender{}, and \varethnicity{} as demographic variables (Table~\ref{tab:demo-var}).
%We use \varage{}, \vargender{},\footnote{We only source \texttt{female} and  \texttt{male} patients. Even though there are instances of other genders in MIMIC-IV data, these are very rare and for that reason not included in our analysis.} and \varethnicity{} as demographic variables (Table~\ref{tab:demo-var}).
We choose these variables for their relevance to studying model biases, their predictive association to our downstream tasks, and their availability~\cite{vanAken2021, celi2022sources, Zhao2024}.
In MIMIC-IV demographic variables are redacted from discharge notes to de-identify protected health information; we thus extract these variables from the patient's structured data. There are no missing demographic variable.

\paragraph{Vital signs}
In Table~\ref{tab:vitals-var} we show the vital sign variables we use in \benchmark{}: \varheartrate{} (\varheartrateacronym{}), \varsystolicbloodpressure{} (\varsystolicbloodpressureacronym{}), \vardiastolicbloodpressure{} (\vardiastolicbloodpressureacronym{}), \varrespiratoryrate{} (\varrespiratoryrateacronym{}), \varoxygensaturation{} (\varoxygensaturationacronym{}),  and \varbodytemperature{} (\varbodytemperatureacronym{}).
These are all clinically relevant indicators in early detection and prognosis across clinical specialties~\cite{downey2017strengths,alghatani2021predicting, herasevich2022impact}, and abundant in MIMIC-IV, having missing values ranging between 4\% and 27\%.
We extract all vital sign variables directly from patients' admission notes using LLaMA-3.3-70B-Instruct with few-shot prompting~\cite{grattafiori2024llama}.
For more details, please refer to \S~\ref{sec:extraction-specifications}.

%\paragraph{Missing values}
%There are no missing demographic variables and
%i.e., missingness is $0\%$.
%vital sign variables have missing values ranging between 4\% and 27\%.

\begin{table*}[!t]
\centering
\resizebox{\textwidth}{!}{
\begin{tabular}{lcccccl}
\toprule
\textbf{Vital Sign} (\textbf{Range} $\mathcal{X}_v$) &
\textbf{Very low (-2)} &
\textbf{Low (-1)} &
\textbf{Normal (0)} &
\textbf{High (+1)} &
\textbf{Very high (+2)} &
\textbf{Missing (\%)} \\
\midrule

\textbf{Heart Rate (1--200 bpm)} &
1--40 &
41--50 &
51--90 &
91--110 &
111--200 &
$5.7\%$ \\

\textbf{Systolic Blood Pressure (1--220 mmHg)} &
1--70  &
71--89  &
90--119  &
120--139  &
140--220 &
$4.3\%$  \\

\textbf{Diastolic Blood Pressure (1--140 mmHg)} &
1--40 &
41--59 &
60--79 &
80--89 &
90--140 &
$4.3\%$  \\

\textbf{Respiration Rate (1--50  bpm)} &
1--8 &
9--11 &
12--20 &
21--24 &
25--50 &
$12\%$ \\

\textbf{Oxygen Saturation (1--100 \%SpO\textsubscript{2})} &
1--91 &
92--95 &
96--100 &
--- &
--- &
$5.1\%$ \\

\textbf{Temperature (70.0--110.0 $^o$F)} &
70.0--89.4 &
89.5--94.9 &
95.0--100.2 &
100.3--103.9 &
104.0--110.0 &
$27\%$ \\

\bottomrule
\end{tabular}}
\caption{Clinically meaningful ranges and severity scores. Severity ranges are derived from clinical guidelines.}
\label{tab:vitals-var}
\end{table*}

\subsection{Clinically-informed data preparation}
\label{sec:data_preparation}
%Below, we explain how we preprocess variables $v \in \mathcal{V}$.
We map numerical variables onto bins and explain how we do this for demographics and vital signs.

%\begin{itemize}
%    \item Demographics $\rightarrow$ ?
%    \item Vital signs $\rightarrow$ severity scales
%    \item Clinical guidelines
%    \item Raw admission notes vs. template-based notes
%\end{itemize}

\paragraph{Demographic variables}
The only numerical demographic variable is \varage{}; \vargender{} and \varethnicity{} are already categorical.
%Moreover, MIMIC includes only adult ICU patients with \varage{} capped at 89 years as part of its pseudonymisation procedure, i.e., any patients with $\varage{} > 89$ are mapped to $\varage{} = 91$. 
We preprocess \varage{} by binning ages into four classes, as illustrated in Table~\ref{tab:demo-var}.
%We preprocess \varage{} by removing any patients with $\varage{} < 18$ or $\varage{} > 91$ years and bin ages into four classes, as illustrated in Table~\ref{tab:demo-var}.

\paragraph{Vital sign variables}
All vital signs are numerical variables and are mapped onto \textit{clinically-grounded severity scales}.
For a given variable $v \in \mathcal{V}$, these scales provide immediate clinical interpretation: \texttt{normal} values indicate an ideal state, \texttt{high} or \texttt{low} values reflect a deterioration in the patient’s condition, and \texttt{very high} or \texttt{very low} values correspond to the most critical states.
These severity scales are derived from established clinical guidelines for each vital sign and discretize continuous measurement into small set of interpretable ranges to enable consistent counterfactual analysis ~\cite{idsa_fever_2003,news2_2017,sccm_temp_2015,who_hypoxemia_2016,sccm_sepsis_2021,aha_american_2024}. Oxygen saturation is a special case, as values above \texttt{normal} are not defined.
%, as the highest saturation is not clinically meaningful. 
For more details refer to \S~\ref{sec:clinical_guidelines}.
%\red{Explain how we map vital signs to severity scales. Mention oxygen saturation not having entries higher than `normal' in the scale and explain in 1 sentence. Mention the guidelines and point the reader to the appendix for more details.}

\subsection{Counterfactual generation}
\label{sec:counterfactuals}
%\begin{itemize}
%    \item Generation
%    \item Validation
%\end{itemize}

%\paragraph{Generation}
We generate counterfactual notes by modifying one clinical variable $v \in \mathcal{V}$ at a time---either a demographic attribute or a vital sign---while keeping the rest of the note unchanged.
For each patient $i \in \mathcal{P}$, we uniformly draw five samples per class per variable $v \in \mathcal{V}$.
Classes for demographic variables are defined in Table~\ref{tab:demo-var} and classes for vital sign variables are the five severity bins in Table~\ref{tab:vitals-var}.
We draw five counterfactuals uniformly from each class to ensure a balanced coverage across all classes. %The expectation is that small physiological perturbations should induce proportionally small prediction changes; large shifts under minor changes indicate poor robustness.
%Because vital signs in text often differ from structured data, we extract vital sign values from the \textit{physical exam} section using few-shot prompting with LLaMA 3 70B~\cite{grattafiori2024llama}.
%Earlier experiments with rule-based extraction proved unreliable due to extensive variability in vital signs documentation.
In total, we generate \textbf{166,731 counterfactuals}: 20,020 (\varage{}), 1,001 (\vargender{}), 4,004 (\varethnicity{}),  30,895 (\varsystolicbloodpressureacronym{}), 30,895 (\vardiastolicbloodpressureacronym{}), 27,070 (\varheartrateacronym{}), 20,590 (\varbodytemperatureacronym{}), 19,404 (\varrespiratoryrateacronym{}), and 12,852 (\varoxygensaturationacronym{}).

\subsection{Automatic and human validation}
\label{sec:data_validation}
We ensure \benchmark{}'s data quality (including counterfactuals) through automatic and manual checks.

\paragraph{Variable extraction and replacement}
As mentioned earlier, demographic variables are extracted from patient structured data, whereas vital sign variables are extracted from admission notes using LLaMA-3.3-70B-Instruct with few-shot prompting. %~\cite{grattafiori2024llama}.
For each variable replacement, we first use the GNU diff tool\footnote{\url{https://www.gnu.org/software/diffutils/}} to verify that edits affect only the intended variable, flagging anomalies for human review.
We manually validated all the raw and template-based notes from original and counterfactual data in \benchmark{} and observed $\sim5\%$ error rate in vitals extracted with LLMs, while demographic replacements drawn from structured data showed no error.
All errors were manually corrected.

\section{Experimental Setup}
\label{sec:experimental_setup}

We now describe how we assess LLMs' sensitivity and robustness to counterfactuals.
Towards this purpose, we describe our evaluation protocol and experiments using two downstream tasks, mortality and length-of-stay prediction  (\S~\ref{sec:task_specific_evaluation}), and also independently of any downstream task (\S~\ref{sec:task_independent_evaluation}).
We provide details on the multiple LLMs in \S~\ref{sec:models}.
%We provide details on the multiple LLMs as well as in which cases we use in-context learning and supervised fine-tuning in \S~\ref{sec:models}.

\subsection{Task-specific evaluation}
\label{sec:task_specific_evaluation}
%We apply the cohort selection criteria below for all splits.

\paragraph{Inclusion and exclusion criteria}
We perform mortality and length-of-stay prediction in the ICU within the first 24 hours.
We only use the first ICU stay for the patients in our cohort and, for patients with multiple ICU stays: when these occur less than 48 hours apart, we merge them into a single ICU stay; when ICU stays are separated by more than 48 hours, we treat them as distinct ICU admissions and keep only the first admission.
We do not include patients who die within the first 24 hours of admission to the ICU or who have an ICU stay shorter than 24 hours.

\subsubsection{Mortality classification}
\label{sec:mortality_prediction}
We frame mortality prediction as binary classification with labels $\mathcal{Y}^\text{mort} \in \{0,1\}$ indicating if the patient dies ($1$) or not ($0$) during the hospital admission.
The mortality rate in our cohort is 7\%.

%\red{Explain how splits are created and how much data they have. Following previous work? Temporal structure? Cite.}

\subsubsection{Length-of-Stay (LoS) classification}
\label{sec:los_prediction}
LoS measures how long a patient stays in the ICU and is defined as the number of hours between admission and discharge.
LoS prediction is modelled as a 4-way classification task with labels $\mathcal{Y}^\text{los} \in \{0,1,2,3\}$ based on training data quantiles:
label 0 for $\texttt{LoS} < Q_{25}$ ($24$ to $39$ hours), 
label 1 for $Q_{25} \leq \texttt{LoS} < Q_{50}$ ($39$ to $59$ hours), 
label 2 for $Q_{50} \leq \texttt{LoS} < Q_{75}$ ($59$ to $112$ hours), and 
label 3 for $\texttt{LoS} \geq Q_{75}$ (over $112$ hours).
%These LoS buckets follow previous work, and are approximately evenly represented, with each class accounting for roughly 25\% of the data ~\cite{hempel2023prediction}.
%The majority class ($XX$ to $YY$ days) has a \red{$YY\%$} ratio whereas the minority class ($XX$ to $YY$ days) has a \red{$XX\%$} ratio.
LoS bins are evenly represented with each class accounting for $\sim$25\% of the data.
 
%\red{Explain how splits are created and how much data they have. Following previous work? Temporal structure? Cite.}

\subsubsection{Evaluation metrics}
We use the following evaluation metrics for mortality (\S~\ref{sec:mortality_prediction}) and length-of-stay  (\S~\ref{sec:los_prediction}) prediction.

\paragraph{Label probability shift}
For all variables $v \in \mathcal{V}$, \texttt{KL} is the KL divergence that measures how distant are label probability distributions for counterfactuals
$\xcounterfactualspervariable$ %$\tilde{\mathcal{X}}^i_j$ 
relative to original patients 
$\mathbf{x}_{iv}$. %$X^i_j$, 
%averaged across all patients.
\begin{align}\label{eq:label-prob-shift}
    %\scalemath{0.75}{
    \texttt{KL} &= \EX_{i \in \mathcal{P}, \ \xcounterfactualspervariable \in  \mathcal{C}_{i}} %\alpha \cdot
            %\sum_{i \in \mathcal{P}}^{} \sum_{  \xcounterfactualspervariable \in  \mathcal{C}_{iv}  }^{}
                \Big[ g(\xcounterfactualspervariable, \mathbf{x}_{iv}) \Big], \\
            %    \big[& \\
            %        \infdiv{ f_\theta(Y | \xcounterfactualspervariable) }
            %               { f_\theta(Y | \mathbf{x}_{iv}) }
            %    \big]&,\\
    %\alpha &= \frac{1}{|\mathcal{P}| * |\mathcal{C}_{iv}|},\\
    g(a,b) &=
                \infdiv{ f_\theta(Y | a) }
                       { f_\theta(Y | b) },\nonumber
    %}
\end{align}
\noindent
where $f_\theta$ is an LLM with parameters $\theta$ that computes the probability distribution over labels $Y$.
%\paragraph{Expected LoS shift}
%For each variable $j$, we compute the expected change in the probability distribution $\Delta P_j$ over labels $Y$ under counterfactuals.
%
%\begin{equation*}
%    \Delta P_j = \EX_{p_i \sim \mathcal{P}}{}\EX_{X^i_j}{[P_{LLM} (Y | \tilde{X}^i_{j'}) - P_{LLM} (Y|X^i_j)]}
%\end{equation*}
%
%For each downstream task, we denote the expected change in probability $\Delta_{LOS}$ for the shift in length-of-stay and $\Delta_{M}$ for the change in mortality between original and counterfactual patients and defined these quantities as below.
%
%\begin{equation*}
%    \Delta_{LOS} = \frac{1}{|\tilde{\mathcal{X}}^i_j|} \sum_{j=1}^{|\tilde{\mathcal{X}}^i_j|}{ \Delta P_j }
%\end{equation*}

\paragraph{Percentage of label flips}
For all variables $v \in \mathcal{V}$, we compute the percentage of counterfactuals which induce a flip in the highest probability label.
This metric measures prediction stability in terms of class labels and indicates whether counterfactual perturbations caused categorical prediction shifts.
\begin{align}\label{eq:label-flips}
    \texttt{Flips(\%)} &=
            \EX_{i \in \mathcal{P}, \ \xcounterfactualspervariable \in  \mathcal{C}_{i}}
            %\alpha \cdot
            %\sum_{i \in \mathcal{P}}^{}{
            %        \sum_{\xcounterfactualspervariable \in  \mathcal{C}_{iv}}^{}{
                            \Big[ h(\xcounterfactualspervariable, \mathbf{x}_{iv}) \Big], \\%\text{f}_{i,j'}
            %        }
            %},\\
    %\alpha &= \frac{100}{|\mathcal{P}| * |\mathcal{C}_{iv}|},\nonumber\\
    h(a,b) &= %\text{f}_{i,j} &=
    \begin{cases}
    1 : \text{argmax}f_\theta(Y | a) \neq \\ \qquad\qquad\text{argmax}f_\theta(Y | b) \nonumber\\
    0 : \text{otherwise}
    \end{cases}
\end{align}
\noindent
where $f_\theta$ is an LLM with parameters $\theta$ that computes the probability distribution over labels $Y$. 

We approximate the expectations in Eqs.~(\ref{eq:label-prob-shift}--\ref{eq:label-flips}) using counterfactual samples described in \S~\ref{sec:counterfactuals}.

\paragraph{Correct direction}
In addition to probability shift and label stability, we assess whether counterfactual perturbations induce prediction changes consistent with the expected direction of clinical severity. 
For a counterfactual $\xcounterfactualspervariable \in \mathcal{C}_{iv}$ for a patient $i$ and variable $v$, we define the severity shift as 
\begin{align}
    \Delta s_{iv} = |\text{sev}(\xcounterfactualspervariable)| - |\text{sev}(\mathbf{x}_i)|,
\end{align}
\noindent
where $\text{sev}(\cdot)$ denotes the severity scale associated with the variable value.
A positive shift ($\Delta s_{iv}>0$) corresponds to increased severity, while a negative shift ($\Delta s_{iv}<0$) corresponds to decreased severity.

We denote the expected LoS for a patient $i$ by
\[
\mathbb{E}[y_i^{\text{los}} \mid \mathbf{x}_i] = \sum_{l \in \mathcal{Y}^{\text{los}}} f_\theta(l \mid \mathbf{x}_i) \cdot \mu_l,
\]
where $\mu_l$ is the empirical mean LoS for bin $l$ computed over the training data.
Similarly, we denote the expected mortality for a patient $i$ simply by
\[
\mathbb{E}[y_i^{\text{mort}} \mid \mathbf{x}_i] =  f_\theta(y^{\text{mort}}_i{=}1 \mid \mathbf{x}_i).
\]
%For mortality, we use the predicted probability of death $f_\theta(y^{\text{mort}}{=}1 \mid \mathbf{x})$.

Finally, we consider downstream predictions for a counterfactual as directionally correct iif
%\[
\begin{align}\label{eq:correct-direction}
    \text{s}\!\left( \mathbb{E}[y_i \mid \xcounterfactualspervariable] - \mathbb{E}[y_i \mid \mathbf{x}_i]) \right)
    =
    \text{s}(\Delta s_{iv}),
\end{align}
%\]
where $y_i \in \{y_i^{\text{los}}, y_i^{\text{mort}}\}$ denotes the expected LoS and mortality probability, respectively, and $\text{s}(\cdot)$ is the sign function.
We report \texttt{\%CD} as the proportion of counterfactuals that satisfy Equation~(\ref{eq:correct-direction}).

\paragraph{Monotonicity}

For each severity scale $s \in \{-2,-1, \;0,+1,+2\}$ (see Table~\ref{tab:vitals-var}), we compute the
%mean difference between the prediction for a counterfactual $\xcounterfactualspervariable$ and its original patient $\mathbf{x}_i$.
mean of the differences between expected downstream predictions for original and counterfactual patients as below.
\begin{align}
    m(s) = \mathbb{E}[y_i \mid \xcounterfactualspervariable] - \mathbb{E}[y_i \mid \mathbf{x}_i],
\end{align}
\noindent
where $y_i \in \{y_i^{\text{los}}, y_i^{\text{mort}}\}$ again denotes the expected LoS and mortality probability, respectively.
We compute $m(s)$ for each $s$ by aggregating all original--counterfactual pairs ($\mathbf{x}_i, \xcounterfactualspervariable$) with a corresponding $\Delta s_{iv} = s$.

We consider predictions as monotonic iff all the inequalities below hold.
\begin{align}
    m(s-1) \le m(s) \le m(s+1), \qquad \text{and}\\
    \text{if} (s \neq 0)
    \begin{cases}
    s \in \{-2, -1\} : m(s) < 0 \nonumber\\
    s \in \{+1, +2\} : m(s) > 0
    \end{cases}
\end{align}
%$m(-2) \le m(-1) \le m(0) \le m(1) \le m(2)$ and $m(-2),m(-1)<0$, $m(1),m(2)>0$.

We report \texttt{\%Mono} as the proportion of predictions satisfying both conditions.

\subsection{Task-independent evaluation}
\label{sec:task_independent_evaluation}
Here we analyze how the probability an LLM assigns to the original patient changes under a counterfactual \textit{independently of any task}.

\subsubsection{Evaluation metrics}
We first note that we use perplexity as a proxy for the probability of an admission note as below.
\begin{equation*}
\textstyle
    \mathrm{PPL} = \exp\!\left( -\frac{1}{N} \sum_{n=1}^{N} \log P(t_n) \right),
\end{equation*}
\noindent
where $t_n$ are the tokens in a patient note.

\paragraph{Patient probability shift}
%To implement the shift in probability brought by a counterfactual relative to an original patient, we directly compute
We calculate the expected shift in perplexity $\Delta\texttt{PPL}$ brought by a counterfactual relative to an original patient as below.
%The patient probability shift is computed as below.
\begin{equation}\label{eq:patient-prob-shift}
    \scalemath{0.87}{
    \Delta\texttt{PPL} = \EX_{i \in \mathcal{P}, \ \xcounterfactualspervariable \in  \mathcal{C}_{iv}} \left[ \mathrm{PPL}_{\xcounterfactualspervariable} - \mathrm{PPL}_{\mathbf{x}_{iv}} \right],
    }
\end{equation}
where $\mathrm{PPL}_{\xcounterfactualspervariable}$ and $\mathrm{PPL}_{\mathbf{x}_{iv}}$ are the perplexities of counterfactual and original patient admission notes, respectively.

We approximate the expectation in Eq.~(\ref{eq:patient-prob-shift}) using counterfactual samples described in \S~\ref{sec:counterfactuals}.

\subsection{Large language models (LLMs)}
\label{sec:models}
We analyze the performance of 8 LLMs via in-context learning (zero-shot)
%and few-shot prompting) 
%and \red{XX} LLMs fine-tuned 
for downstream tasks.
For our experiments, we choose LLMs of varying sizes, architectures, domains, and reasoning capabilities: \textbf{Phi4-14B}~\cite{abdin2024phi}, \textbf{Meditron3-Phi4-14B}~\cite{meditron3phi4}, \textbf{LLaMA-3.3-Instruct-70B}~\cite{grattafiori2024llama}, \textbf{OpenBioLLM-70B}~\cite{OpenBioLLMs}, \textbf{DeepSeek-R1-Distill-70B}~\cite{guo2025deepseek}, \textbf{Qwen-2.5-Instruct-72B}~\cite{qwen2.5}, \textbf{GPT-OSS-120B}~\cite{openai2025gptoss120bgptoss20bmodel}, and \textbf{GPT-4.1-mini}~\cite{openai_gpt41mini}.\footnote{We test GPT 4.1 mini within a private and secure environment according to the MIMIC-IV data usage agreement.}
Zero-shot 
%and few-shot 
prompts we use are available in \S~\ref{app:zero-shot-prompts}.
%When applying few-shot learning, we randomly sample 1--3 demonstrations (input--label pairs) from the validation set and reuse the same demonstrations across all LLMs.

%\paragraph{Fine-tuned LLMs}
%We use OpenBioLLM~\cite{OpenBioLLMs} (70B), Meditron3-Phi4~\cite{meditron3phi4} (14B), Phi-4~\cite{abdin2024phi} (14B), LLaMA 3.1 Instruct~\cite{grattafiori2024llama} (70B), \red{GPT-OSS [CITATION NEEDED] (SIZE NEEDED)}, and DeepSeek-R1-Distill~\cite{guo2025deepseek} (70B).
%LLMs are fine-tuned for mortality prediction (\S~\ref{sec:mortality_prediction}) and length-of-stay prediction (\S~\ref{sec:los_prediction}).

We analyze LLMs along three axes: 
(1) \textit{template vs. raw admission notes}
(2) \textit{task-independent vs. task-specific}
(3) \textit{medical-domain vs. general-purpose vs. reasoning-oriented}.
These comparisons reveal how model specialization and reasoning capacity affect robustness, sensitivity, and behavioral consistency. We summarise all LLM specifications in Table~\ref{tab:llm-models}, \S~\ref{app:llm-models}.
%This selection is not intended to be exhaustive, but captures diversity across base architecture, size, domain, and reasoning focus, offering a representative analysis of current LLM behavior.
%\textit{We do not use proprietary LLMs since the data usage agreement of MIMIC-IV does not allow for it.}\footnote{\url{https://physionet.org/about/licenses/physionet-credentialed-health-data-license-150/}}

\section{Results}

Below we report results for our task-independent (\S~\ref{sec:results-task-independent}) and task-specific  (\S~\ref{sec:results-task-specific}) evaluations.

\subsection{Task-independent}\label{sec:results-task-independent}
\begin{figure*}[!t]
    \centering
    \begin{subfigure}[b]{0.48\textwidth}
        \centering
        \includegraphics[width=\linewidth]{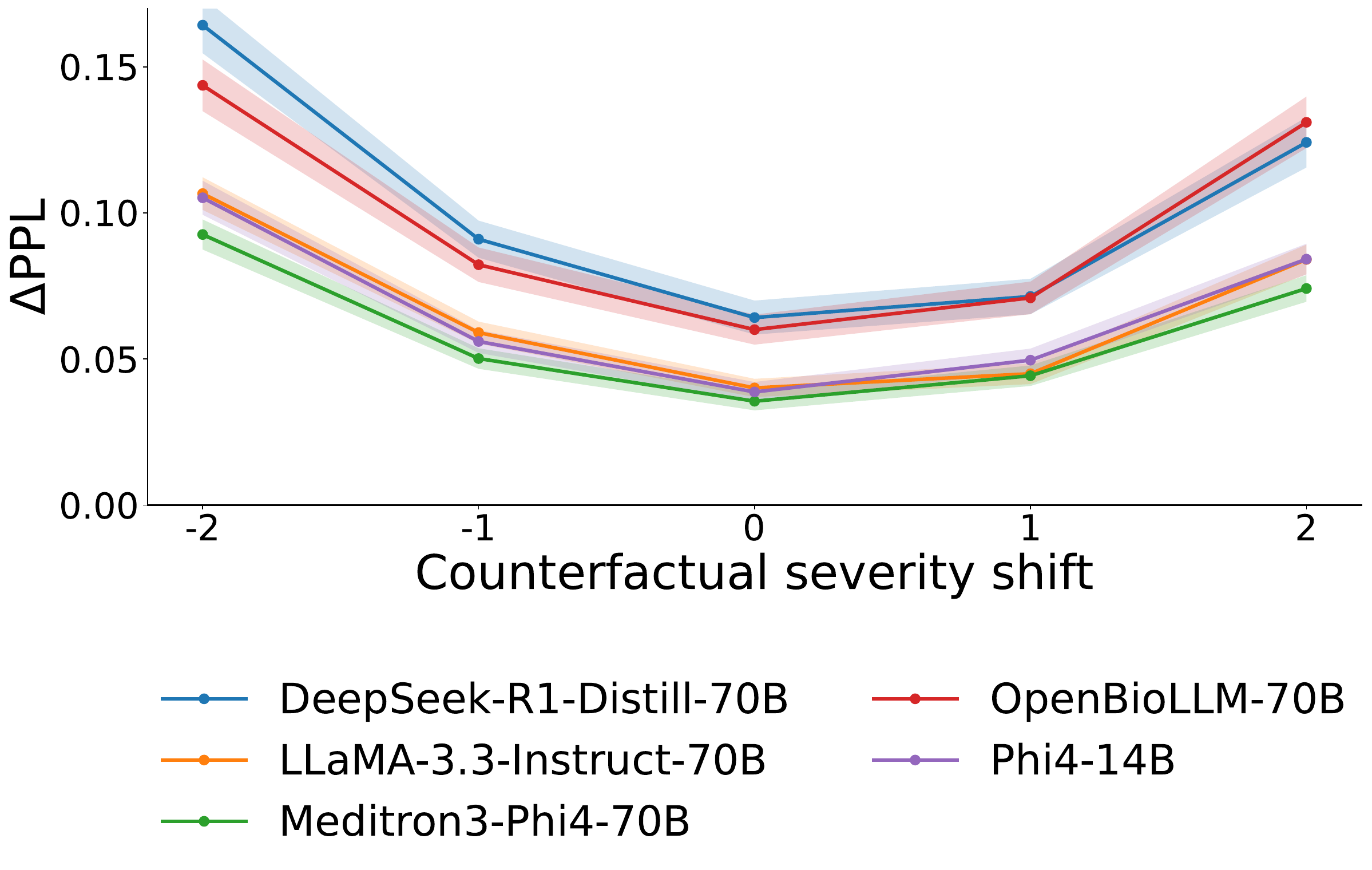}
        \caption{Raw notes.}
        \label{fig:notask-r}
    \end{subfigure}
    \hfill
    \begin{subfigure}[b]{0.48\textwidth}
        \centering
        \includegraphics[width=\linewidth]{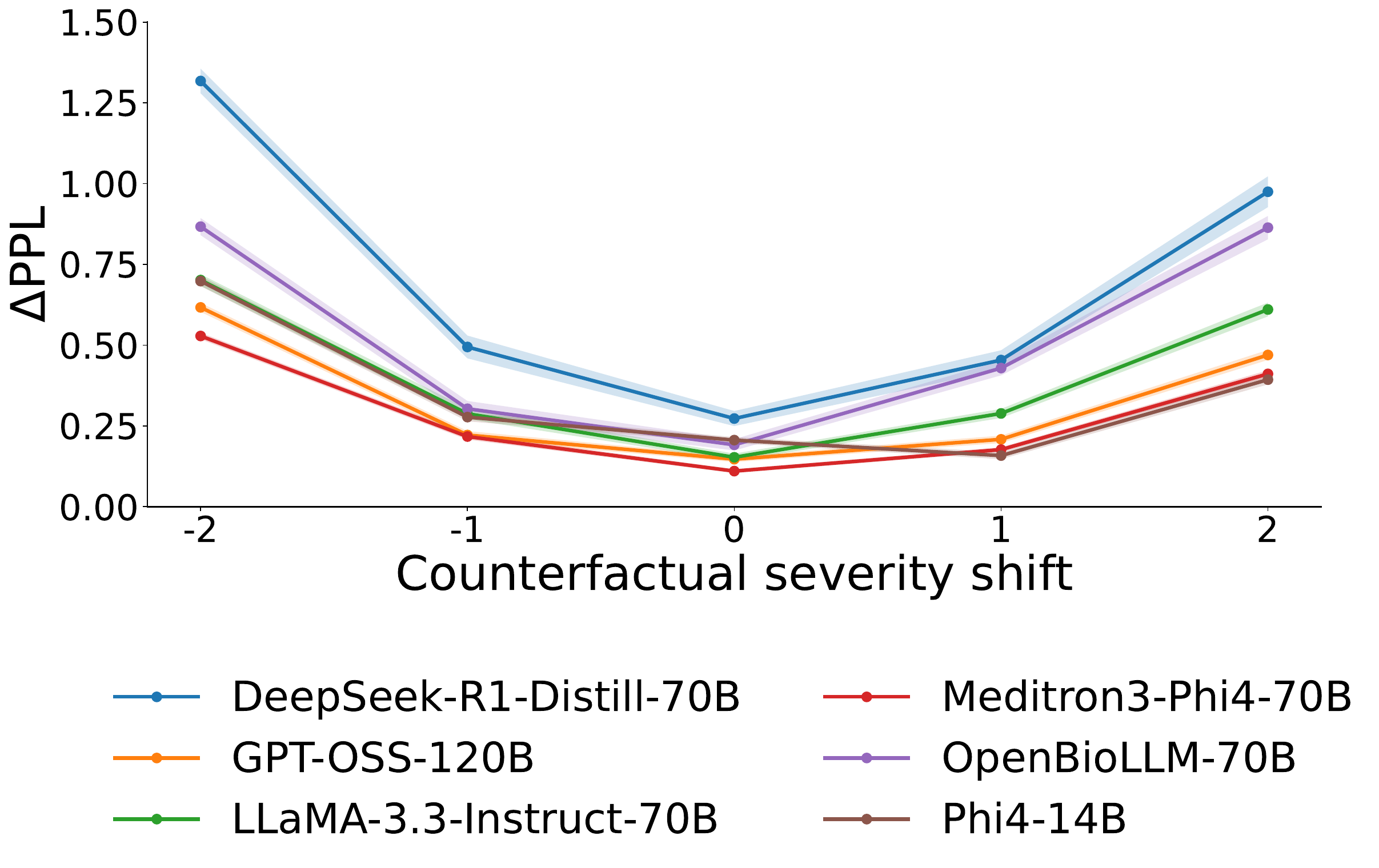}
        \caption{Template-based notes.}
        \label{fig:notask-t}
    \end{subfigure}
    \caption{Average per-token $\Delta$\texttt{PPL} across counterfactual severity shifts. $\Delta\texttt{PPL}$ grows with both increasing and decreasing severity, indicating consistent linguistic sensitivity. In (a) \gptoss was excluded due to its substantially higher $\Delta\texttt{PPL}$ ($2.5\pm5.1$). We observe the same pattern in (a) and (b) with higher effects in template notes. Except \gptoss that inversely, presents a smaller response in raw notes. \gptmini is omitted because we could not obtain per-token log probabilities for this model.}
    \label{fig:notask}
\end{figure*}

%\subsection{Vital Signs}
%\parahead{Task-independent} 
We observe small behavioral differences 
%already emerge without an explicit prediction task. Changes in perplexity remain 
in terms of change in perplexity across models that scale scale with severity (Figure~\ref{fig:notask}). These differences are consistent in both note modalities, being effect on template-based notes being more pronounced than in raw notes. 
This means that models are linguistically \textit{surprised} by counterfactuals even when not trained for a downstream task. \gptoss shows a substantially larger $\Delta$\texttt{PPL} than the other models, suggesting heightened sensitivity or instability to counterfactual perturbations.
\begin{comment}
\begin{figure}[!t]
    \centering
    \includegraphics[width=\linewidth]{notask.pdf}
    \caption{Average per-token $\Delta$\texttt{PPL} across counterfactual severity shifts (raw notes).
    %Positive severity shifts denote models find counterfactuals more severe than original value; negative severity shift implies the opposite.
    $\Delta\texttt{PPL}$ grows with both increasing and decreasing severity, indicating consistent linguistic sensitivity. \gptoss was excluded for clarity due to its substantially higher $\Delta\texttt{PPL}$ ($2.5\pm5.1$). \gptmini is omitted because we could not obtain per-token log probabilities for this model.}
    \label{fig:notask}
\end{figure}
\end{comment}
\subsection{Task-specific}\label{sec:results-task-specific}
\begin{figure}[!t]
    \centering
    \includegraphics[width=\linewidth]{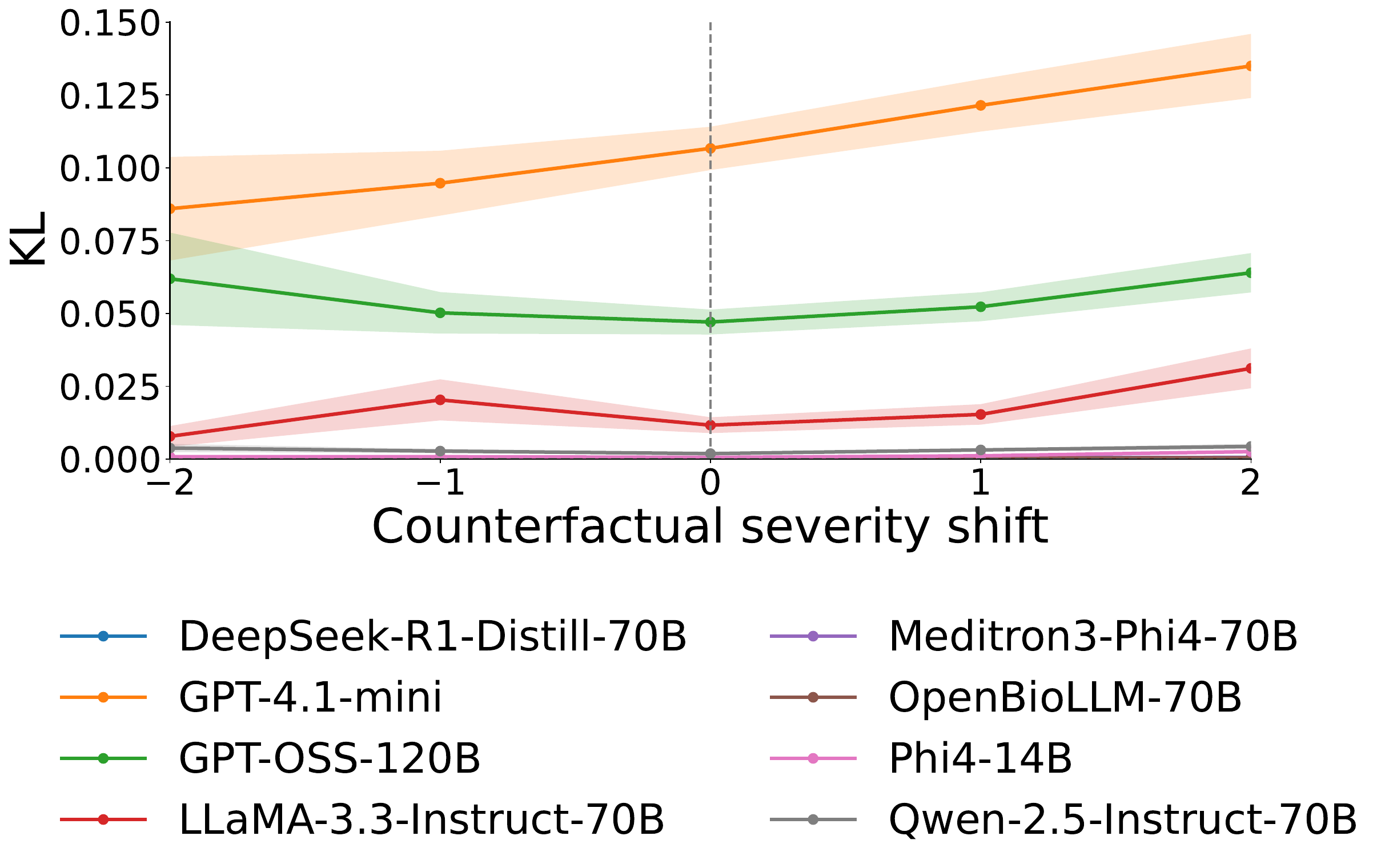}
    \caption{Average \texttt{KL} across counterfactual severity shifts for LoS task (raw notes). Models show different behaviours. \gptmini, \gptoss and \llama show larger KL shifts than other models.}
    \label{fig:kl-los}
\end{figure}

\parahead{Vital signs}
We observe behavioral differences in the presence of downstream tasks (Figure~\ref{fig:kl-los}).
As with perplexity, \texttt{KL} also often scales with severity shifts in several models. However, the behaviors are more heterogeneous, \gptmini presenting a monotonic increase, and \meditron and \obllm almost flat response.
Moreover, \gptoss again exhibits substantially a larger KL shift than the other models.
%Performance of standard machine learning models (Table~\ref{tab:baselines}) and LLMs are comparable (Table~\ref{tab:model_summaries}) for both tasks.
LLM performance on template-based data, in both tasks, is comparable to standard machine learning models (i.e., see Table~\ref{tab:model_summaries} and Table~\ref{tab:baselines} in \S~\ref{app:additional-results}).
For mortality prediction, class imbalance impacts the performance of both classical machine learning and LLM-based models.
\begin{table}[t!]
\centering
{\resizebox{\columnwidth}{!}{%

\begin{tabular}{r r r r r r r} 
\toprule
\textbf{Model} &
\textbf{Acc.} & \textbf{F1} &
\textbf{$\Delta$KL$\pm$std} &
\textbf{\%CD} & \textbf{\%Flip} & \textbf{\%Mono} \\
\midrule

% ===================== Mortality -- raw =====================
\multicolumn{7}{l}{\textbf{Mortality prediction, raw notes}} \\
\midrule
\phiF & \underline{0.28} & \underline{0.26} & \underline{0.00} $\pm$ 0.01 & \textbf{73.1} & 0.9 & \textbf{100.0} \\
\rowcolor{medcolor}
\meditron & 0.41 & 0.36 & \underline{0.00} $\pm$ \underline{0.00} & 69.8 & \underline{0.6} & \textbf{100.0} \\
\rowcolor{reasoncolor}
\deepseek & 0.60 & 0.48 & \underline{0.00} $\pm$ 0.01 & 72.4 & 1.8 & \textbf{100.0} \\
\llama & 0.71 & 0.54 & \textbf{0.97} $\pm$ \textbf{2.31} & 69.5 & \textbf{15.0} & \underline{62.5} \\
\rowcolor{medcolor}
\obllm & 0.82 & 0.62 & 0.01 $\pm$ 0.02 & \underline{54.8} & 4.0 & 75.0 \\
\qwen & 0.67 & 0.52 & 0.01 $\pm$ 0.04 & 69.7 & 1.4 & \textbf{100.0} \\
\gptoss & \textbf{0.88} & \textbf{0.63} & 0.06 $\pm$ 0.12 & 64.6 & 4.1 & \textbf{100.0} \\
\gptmini & 0.62 & 0.50 & 0.18 $\pm$ 0.32 & 62.9 & 7.1 & 87.5 \\
\cmidrule{2-7}
%\rowcolor{black!6}
\textbf{Average} & 0.62 & 0.49 & 0.15 $\pm$ 0.35 & 67.1 & 4.4 & 90.6 \\
\midrule

% ===================== Mortality -- template =====================
\multicolumn{7}{l}{\textbf{Mortality prediction, template-based notes}} \\
\midrule
\phiF & \underline{0.73} & 0.53 & 1.37 $\pm$ 1.07 & 90.1 & 32.4 & \textbf{100.0} \\
\rowcolor{medcolor}
\meditron & 0.84 & 0.57 & 0.22 $\pm$ 0.14 & \textbf{90.9} & 26.6 & \textbf{100.0} \\
\rowcolor{reasoncolor}
\deepseek & 0.81 & 0.54 & 0.36 $\pm$ 0.15 & 90.7 & 25.0 & \textbf{100.0} \\
\llama & 0.78 & 0.53 & 3.41 $\pm$ 1.93 & 88.3 & 29.5 & 87.5 \\
\rowcolor{medcolor}
\obllm & 0.92 & 0.58 & \underline{0.17} $\pm$ \underline{0.08} & \underline{82.4} & 10.4 & 87.5 \\
\qwen & 0.89 & \textbf{0.59} & 1.11 $\pm$ 0.50 & 89.7 & 16.3 & \textbf{100.0} \\
\gptoss & \textbf{0.93} & \underline{0.48} & 0.21 $\pm$ 0.24 & \underline{82.4} & \underline{1.6} & \underline{75.0} \\
\gptmini & 0.83 & 0.56 & \textbf{3.86} $\pm$ \textbf{2.43} & 88.3 & \textbf{33.3} & 87.5 \\
\cmidrule{2-7}
%\rowcolor{black!6}
\textbf{Average} & 0.84 & 0.55 & 1.34 $\pm$ 0.82 & 87.9 & 21.9 & 92.2 \\
\midrule

% ===================== LoS -- raw =====================
\multicolumn{7}{l}{\textbf{Length-of-stay prediction, raw notes}} \\
\midrule
\phiF & \underline{0.26} & 0.19 & \underline{0.00} $\pm$ 0.01 & \textbf{71.1} & 1.6 & \textbf{100.0} \\
\rowcolor{medcolor}
\meditron & 0.27 & 0.21 & \underline{0.00} $\pm$ \underline{0.00} & 69.8 & 1.9 & \textbf{100.0} \\
\rowcolor{reasoncolor}
\deepseek & 0.28 & \underline{0.17} & \underline{0.00} $\pm$ \underline{0.00} & 68.4 & \underline{0.8} & \textbf{100.0} \\
\llama & \textbf{0.31} & \textbf{0.28} & 0.02 $\pm$ 0.07 & 70.5 & 2.3 & \underline{87.5} \\
\rowcolor{medcolor}
\obllm & 0.29 & 0.25 & \underline{0.00} $\pm$ \underline{0.00} & 62.5 & 2.0 & \underline{87.5} \\
\qwen & 0.30 & 0.21 & \underline{0.00} $\pm$ 0.01 & 65.6 & 1.3 & \textbf{100.0} \\
\gptoss & 0.28 & 0.19 & 0.06 $\pm$ 0.13 & 57.6 & 6.1 & \textbf{100.0} \\
\gptmini & 0.30 & 0.25 & \textbf{0.12} $\pm$ \textbf{0.23} & \underline{56.2} & \textbf{11.3} & \underline{87.5} \\
\cmidrule{2-7}
%\rowcolor{black!6}
\textbf{Average} & 0.29 & 0.22 & 0.03 $\pm$ 0.06 & 65.2 & 3.4 & 95.3 \\
\midrule

% ===================== LoS -- template =====================
\multicolumn{7}{l}{\textbf{Length-of-stay prediction, template-based notes}} \\
\midrule
\phiF & \underline{0.23} & 0.19 & 0.62 $\pm$ 0.43 & 89.5 & 36.9 & \textbf{100.0} \\
\rowcolor{medcolor}
\meditron & \underline{0.23} & 0.20 & 0.13 $\pm$ 0.06 & \textbf{91.2} & 38.1 & \textbf{100.0} \\
\rowcolor{reasoncolor}
\deepseek & 0.27 & 0.16 & 0.03 $\pm$ 0.02 & 77.5 & \underline{9.1} & \textbf{100.0} \\
\llama & 0.26 & 0.17 & 3.23 $\pm$ 1.53 & 88.3 & 32.9 & 87.5 \\
\rowcolor{medcolor}
\obllm & 0.25 & \underline{0.15} & \underline{0.02} $\pm$ \underline{0.01} & 81.5 & 19.0 & 87.5 \\
\qwen & \textbf{0.28} & \textbf{0.22} & 0.87 $\pm$ 0.44 & 81.5 & 29.3 & \textbf{100.0} \\
\gptoss & \textbf{0.28} & 0.17 & 0.51 $\pm$ 0.24 & \underline{51.2} & 27.2 & \underline{75.0} \\
\gptmini & 0.24 & 0.17 & \textbf{3.49} $\pm$ \textbf{1.59} & 86.9 & \textbf{39.8} & 87.5 \\
\cmidrule{2-7}
%\rowcolor{black!6}
\textbf{Average} & 0.26 & 0.18 & 1.11 $\pm$ 0.54 & 81.0 & 29.0 & 92.2 \\
\bottomrule
\end{tabular}}}
\caption{Performance and behavioral metrics across note types and tasks (excluding precision/recall; see Table~\ref{tab:model_summaries_pr}). Orange: medical LLM; blue: reasoning-oriented; white: general-purpose; gray: averages. Highest values are bolded and lowest underlined within each block.}
\label{tab:model_summaries}
\end{table}
%\begin{table}[t!]
%\centering
%\resizebox{.48\textwidth}{!}{
%\begin{tabular}{lcccc}
%\toprule
%\textbf{Model} & \textbf{Acc} & \textbf{F1} & \textbf{Prec} & \textbf{Rec} \\
%\midrule
%logistic (mort)       & 0.65 & 0.21 & 0.13 & 0.66 \\
%xgboost (mort)        & 0.74 & 0.22 & 0.14 & 0.52 \\
%random forest (mort)  & 0.93 & 0.00 & 0.00 & 0.00 \\
%\midrule
%xgboost (los)         & 0.29 & 0.29 & 0.30 & 0.29 \\
%random forest (los)   & 0.28 & 0.28 & 0.28 & 0.28 \\
%\bottomrule
%\end{tabular}}
%\caption{Baseline performance for ICU mortality (mort) and length-of-stay (los) prediction.}
%\label{tab:baselines}
%\end{table}
Overall, we note that using templates
1) improves accuracy,
2) increases the percentage of correct directional responses across both mortality and LoS tasks (Table~\ref{tab:model_summaries}), and
3) amplifies behavioral sensitivity in terms of \texttt{KL} and categorical prediction shifts. 
Medical and reasoning models often show lower \texttt{KL}, suggesting a more conservative overall behavior.
Moreover, general-purpose models achieve higher task performance (especially on mortality) and exhibit greater sensitivity.
\gptoss shows consistently better performance and higher sensitivity to perturbations.  
%Despite the modest effect of perturbations on most of the models in the raw setting, changes are not random. Models consistently exhibit high monotonicity and correct directional behavior, with increased severity leading to higher mortality probability and longer expected LOS, and decreased severity producing the opposite effect (Table~\ref{tab:model_summaries}, Figure~\ref{fig:elos&pmort})
\begin{figure*}[!t]
    \centering
    \begin{subfigure}[b]{0.48\textwidth}
        \centering
        \includegraphics[width=\linewidth]{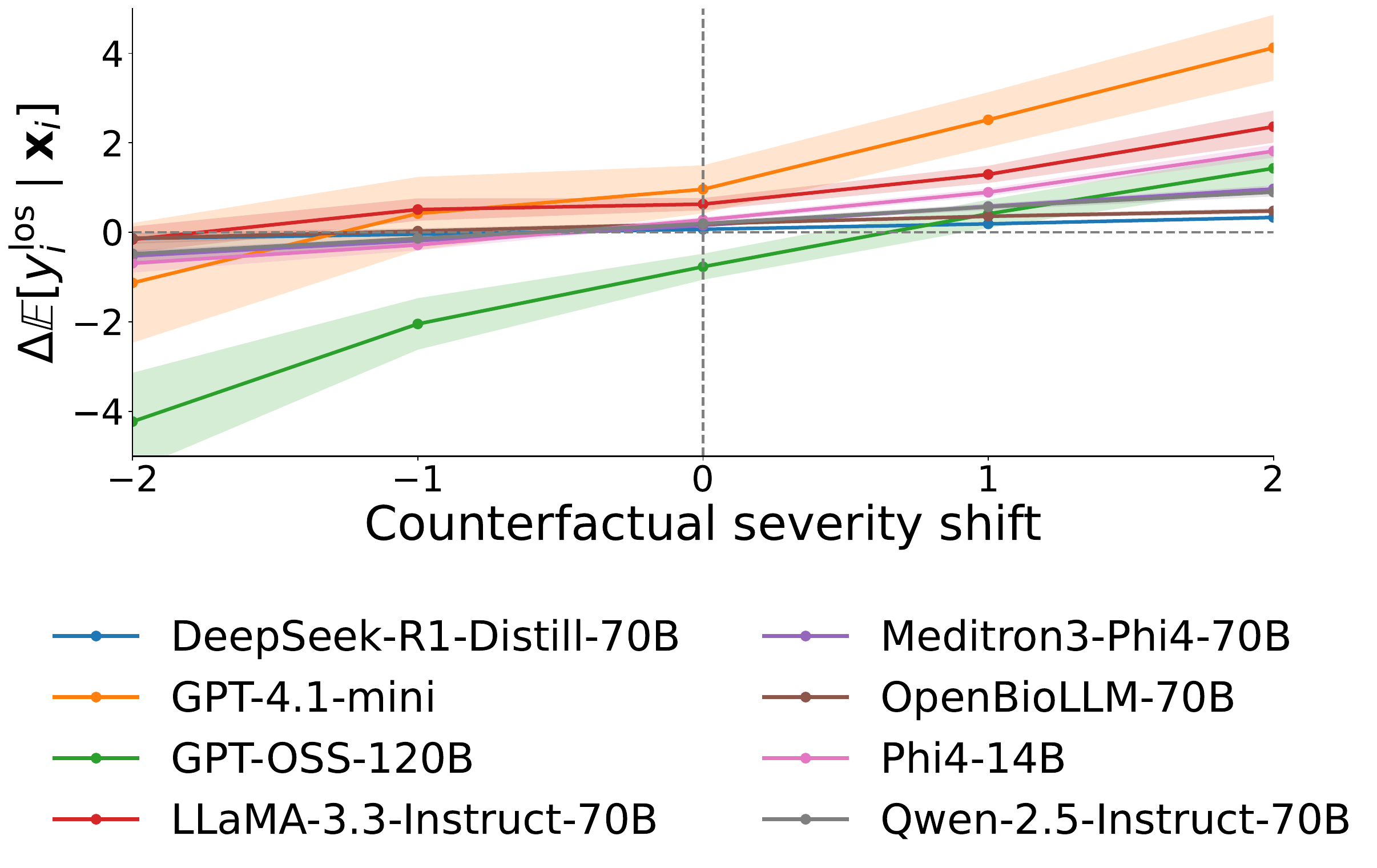}
        \caption{Length-of-stay (LoS).}
        \label{fig:de_los}
    \end{subfigure}
    \hfill
    \begin{subfigure}[b]{0.48\textwidth}
        \centering
        \includegraphics[width=\linewidth]{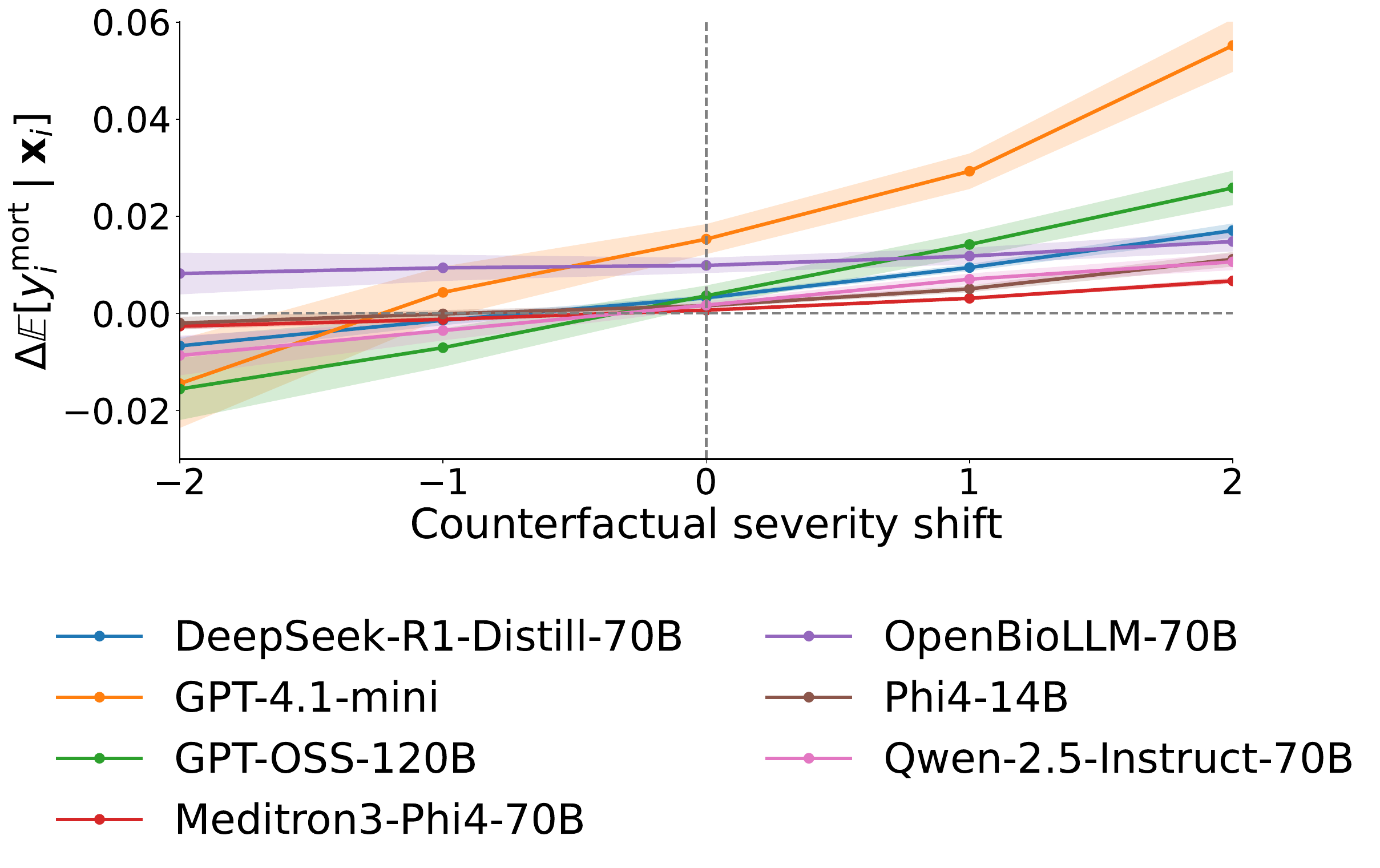}
        \caption{Mortality.}
        \label{fig:dp_mort}
    \end{subfigure}
    \caption{Expected $\Delta \mathbb{E}(y_i^\text{los}|\cdot)$ in hours (a) and probability of mortality $\Delta \mathbb{E}(y_i^\text{mort}|\cdot)$ (b) as a function of counterfactual severity shift. Positive severity shifts are expected to increase predicted LoS and mortality risk, while negative shifts are expected to decrease them. Most models follow a monotonic trend, indicating clinically aligned responses to vital sign counterfactuals. \llama was omitted due to its substantially larger response that dominates the scale, see \S~\ref{app:additional-results}.}
    \label{fig:elos&pmort}
\end{figure*}

\begin{figure*}[!t]
    \centering
    \includegraphics[
        width=\textwidth,
        height=0.30\textheight,
        keepaspectratio
    ]{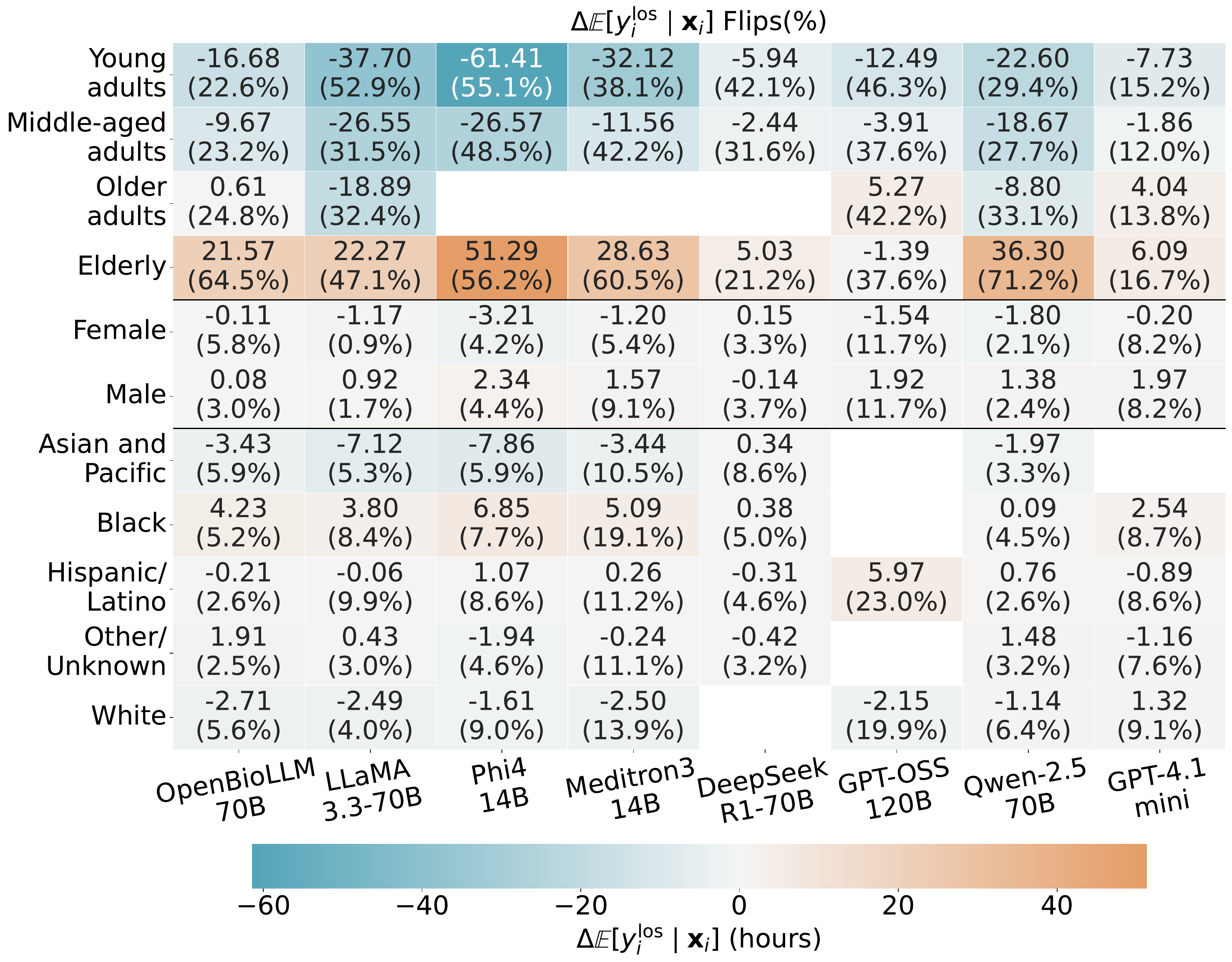}
    \caption{$\Delta \mathbb{E}(y_i^\text{los}|\cdot)$ and \texttt{Flips(\%)} by demographics. Age produces the largest and most significant shifts, while race and gender effects are smaller but consistent. Blank cells denote non-significant t-test comparisons (p<0.05).}
    \label{fig:heatmap_los}
\end{figure*}

\parahead{Demographics}
Demographic perturbations also yield consistent, statistically significant effects (Figure~\ref{fig:heatmap_los}). As expected, age shows the strongest influence: older age groups are associated with longer predicted LoS across models, even under minimal textual edits.

Gender and race effects are smaller on average but still affect boundary decisions, with flip rates \texttt{Flips(\%)} up to 23\%. Although most demographic comparisons are statistically significant, ``Older adults'' age for several models, and ``White'' for \deepseek, and ``Other/Unknown'', ``Black'' and ``Asian and Pacific'' races for \gptoss, do not differ significantly from the remaining demographic groups.

Across models, \deepseek and \qwen remain the most robust models, showing the lowest $\Delta \mathbb{E}(y_i^\text{los}|\cdot)$ and smaller flip rates. In contrast, \phiF and \gptoss produce the largest shifts for race and gender. Race-related effects are slightly larger on average than gender effects, and when significant, most models predict shorter LoS for female, Asian and Pacific, and White patients, and longer LoS for Black patients, aligning with known healthcare disparities~\citep{Macias-Konstantopoulos2023-xi}.
%\begin{figure}[!t]
%    \centering
%    \includegraphics[width=\columnwidth]{heatmap-los.pdf}
%    \caption{Mean $\Delta \mathbb{E}_\text{LOS}$ and flip rates (\%) by demographic class for zero-shot models. Age produces the largest and most significant shifts, while race and gender effects are smaller but consistent. Blank cells denote non-significant comparisons.}
%    \label{fig:heatmap_los}
%\end{figure}

\section{Related Work}
\setlength{\parskip}{0pt}
LLMs are increasingly used in clinical settings, motivating a growing body of work on their robustness, fairness, and interpretability.
\citet{lee2025investigating} analyzed model robustness to distribution shifts and missing data in patient triage, finding that while LLMs outperform traditional models, they exhibit demographic biases.
%In the context of demographic variables, 
\citet{vanAken2021} similarly showed that models with comparable AUROC scores can behave differently when evaluated for sensitivity to
age, gender, and ethnicity.
\citet{Zack2024} and \citet{Zhao2024} further demonstrated that GPT-4 and other LLMs often favor majority groups and yield less accurate predictions for minorities, reinforcing existing disparities.

Beyond fairness, several studies probe robustness through controlled input perturbations.
\citet{MacPhail2024} introduced template-based tests for adverse drug event classification, revealing inconsistent performance even among models with similar discrimination. \citet{Kougia2024} showed that biomedical LLMs frequently fail at ordering events, compromising reliability in decision support.
%Natural Language Inference has been used to assess reasoning consistency.
%, which involves determining the logical relationship between a premise and a hypothesis, has been another focal point in evaluating reasoning capabilities in LLM. 
Likewise, \citet{Altinok2024} and \citet{Aguiar2024} employed natural Language inference-based consistency and faithfulness, uncovering substantial variance across closely related models. 

While these studies highlight the limits of traditional evaluation, most rely on structured templates, leaving raw clinical text largely unexamined.
%Behavioral analysis over raw clinical notes, especially with controlled perturbations, remains largely unexplored.
\textit{Behavioral testing} offers a complementary approach, constructing minimally differing counterfactuals and checking the consistency of model input-output responses ~\cite{beizer1996black}. Originally proposed in software engineering, it has since been adapted to NLP~\citep{ribeiro2020beyond}, vision-and-language~\citep{parcalabescu-etal-2022-valse}, and video-and-language tasks~\citep{kesen2023vilma}. 
%\textit{Behavioral testing (BT)} offers a complementary approach that assesses system behavior without white-box access. BT constructs minimally differing counterfactuals and checks the consistency of model input-output responses~\cite{beizer1996black}. Originally proposed in software engineering, it has since been adapted to NLP~\citep{ribeiro2020beyond}, vision-and-language~\citep{parcalabescu-etal-2022-valse}, and video-and-language tasks~\citep{kesen2023vilma}. 

%Building on this paradigm, we introduce \textbf{\benchmark}, a behavioral evaluation framework that probes model sensitivity to demographic and vital-sign variations in both raw and template-based clinical notes, bridging clinical reasoning analysis and fairness auditing.

\section{Discussion}
\parahead{Task-specific vs. Task-independent} Even without task supervision, models respond systematically to increasing counterfactual severity. Perplexity shifts increase according to severity levels (Fig.~\ref{fig:notask}), indicating that models internally register perturbations as increasingly improbable language events. This supports the use of task-agnostic behavioral probes as an early diagnostic signal of reasoning sensitivity.

\parahead{Template-based vs. Raw notes} 
Templates amplify sensitivity to vital sign changes (Table \ref{tab:model_summaries}), as shown in higher \texttt{KL}s---e.g., 0.15 (raw) $\rightarrow$ 1.34 (templates)---and \texttt{Flips(\%)}, e.g., 4.4$\rightarrow$21.9\% (mortality) and 3.4$\rightarrow$29.0\% (LoS).
On average, templates increase correct directionality of predictions \texttt{\%CD} (67.1$\rightarrow$87.5\% mortality, 66.5$\rightarrow$79.9\% LoS).
We hypothesize that the absence of contextual information forces the model to rely more explicitly on vital signs.
In contrast, raw notes with contextualized reasoning causes importance of vitals to be diluted by other information, potentially under-weighting important physiological changes.
%Variability across both formats, especially for the most reactive models (see Table~\ref{tab:model_summaries}), highlights uncertainty in model behavior to perturbations.

\parahead{Medical vs. General vs. Reasoning LLMs}
Medical models show minimal sensitivity when evaluated on raw notes, likely reflecting conservative priors learned from biomedical corpora, and a preference for holistic interpretation of full clinical notes. Their consistently high \texttt{\%Mono} indicates close alignment with clinical expectations.
When applied to template-based notes their stability decrease and we observe higher \texttt{Flips(\%)}, e.g., \meditron achieves 38.1\% for LoS.

General-purpose models show high sensitivity, which amplifies further under templates. However, this is not consistently accompanied by correct directional or monotonic reasoning, revealing instability despite competitive accuracy. 

\deepseek shows a modest sensitivity compared to medical models even in template settings, while maintaining high directionality and monotonicity.
It favors smaller but calibrated shifts, consistent with expected physiological reasoning. 

\parahead{Demographic Variables}
Demographic attributes subtly but consistently affect predictions. While mean LoS shifts for gender and race were small, elevated flip rates near decision boundaries (Fig.~\ref{fig:heatmap_los}) raise fairness concerns. Consistent prediction patterns across gender and races for most models suggest that pretraining biases are pervasive, reinforcing the need for demographic auditing in clinical LLM evaluation.
\section{Conclusion}
In this work, we introduce \benchmark{} with clinically meaningful counterfactuals.
We show that the behavioral testing of vital signs and demographic variables with counterfactuals provides insights into LLM behavior not captured by standard metrics.
%following previous work findings, we find that models with similar accuracy and F1 scores can respond very differently to minimal, 
Vital sign perturbations consistently induce monotonic and directionally correct shifts in downstream predictions, indicating that models internalize physiological severity, but with substantial variation in sensitivity and stability across architectures. Demographic perturbations produce smaller yet systematic effects, even near decision boundaries, revealing persistent biases that align with known healthcare disparities. 

%Because clinical decisions depend on how risk estimates change as patient conditions evolve, such behavioral differences are clinically relevant, even when aggregate task performance is comparable.
Overall, our results highlight the importance of evaluating not only task performance, but how LLMs respond to clinically meaningful changes in patient information.
We believe \benchmark{} is a step toward more robust, interpretable, and fairness-aware evaluation of clinical LLMs. 

%Our results show the usefulness of examining how medical LLMs respond to clinically-informed counterfactual input changes to understand LLM behavior and reasoning patterns. Our findings reveal that zero-shot models demonstrate coherent reasoning patterns, whereas fine-tuned models are more conservative and show more stable outputs, sometimes even being insensitive to clinically significant changes in the input. Demographic variables affect predictions in a subtle but consistent way, underscoring the need for fairness-aware evaluations. Future research should include fine-tuning strategies that help maintain reasoning capacity and improve calibration and fairness, as well as developing and implementing diagnostic tools to detect and mitigate model bias in real-world clinical settings.

\section*{Limitations} 
%In the following paragraphs, we discuss some limitations of our experimental setup.

\parahead{Datasets and languages}
We use MIMIC-IV, a large publicly available dataset including patients from a single hospital in the US.
We see our work as an important step towards a clinically grounded behavioural testing of medical LLMs.
We believe that important future work lies in further validating the generalisability of our framework in terms of other languages and healthcare systems.

\parahead{Clinical variables}
In this work, we do not cover an extensive range of model capabilities and focus instead on demographic and vital sign variables, focusing on demographic biases and assessing numerical reasoning. We leave other relevant aspects, such as evaluating the temporal reasoning of LLMs, to future work.

%\parahead{Downstream tasks}
%The fine-tuning data for ICU length-of-stay (LOS) prediction was imbalanced, which may have affected model performance. 
\parahead{Missingness in clinical data}
Clinical data is inherently affected by missingness, where absent measurements may reflect clinical practice. Although our analysis includes records with missing values and thus reflects realistic data conditions, explicitly assessing the impact of missingness versus observed inputs on model behavior was beyond the scope of this work.

\parahead{Severity shifts and counterfactual sampling}
Severity shifts are treated symmetrically in our analysis, such that very low and very high values correspond to the same absolute shift (+2), despite potentially differing clinical effects (e.g., on length of stay). Additionally, counterfactual values are sampled uniformly within severity bins, which can yield rare values in extreme bins. Future work should explore alternative sampling strategies better aligned with empirical distributions.
%We used different clinical guidelines to create severity scores for different vital sign variables.
%We tried as much as possible to stick to the nomenclature in the clinical guidelines to emphasise clinical relevance.
%That means that different vital signs can have severity classes with different severity scores, e.g., ``high'' blood pressure is +2, whereas ``high'' temperature is +1.
%We note that this does not affect our analysis, however, since we have conducted only single-variable counterfactuals.
%We believe future work should look into employing clinicians from different clinical specialties to align the severity classes and scores for different clinical variables.

\parahead{Number of counterfactuals}
Some vital signs, such as oxygen saturation and respiration rate, had limited value ranges, resulting in fewer than five unique counterfactuals due to the small number of non-redundant available values.

%\parahead{Clinical validation}
%We did not have a clinical validation of our generated counterfactuals with clinicians.
%We believe an important next step involves including clinicians from different clinical specialties to validate 
%Despite these limitations, our proposed framework remains broadly relevant. The use of MIMIC-IV ensures realism in clinical data, while the counterfactual design enables a controlled setting to evaluate reasoning patterns that can be applied across healthcare contexts. By evaluating a set of architectures, domains and training settings, although not comprehensive, we still capture behavioral trends not tied to a single model. The framework highlights biases and reasoning gaps that are fundamental to understanding safe deployments of medical LLMs, offering contributions that extend beyond the specific dataset and task examined here.  

Despite these limitations, our proposed framework is relevant and we see it as the first step towards clinically-grounded evaluation of LLMs for healthcare.
While not exhaustive, our framework highlights biases and reasoning gaps paving the way to safe deployment of medical LLMs. %, offering contributions that extend beyond the specific dataset and task examined here.  

% commented out for anonymity
\section*{Acknowledgments}
HOB and IC are funded by the project CaRe-NLP with file number NGF.1607.22.014 of the research programme AiNed Fellowship Grants which is (partly) financed by the Dutch Research Council (NWO). EE is funded by the project TUBITAK 2247-A National Outstanding Researchers Program Award No. 123C542.

%\bibliography{mybibliography}

\begin{thebibliography}{45}
\providecommand{\natexlab}[1]{#1}

\bibitem[{Abdin et~al.(2024)Abdin, Aneja, Behl, Bubeck, Eldan, Gunasekar, Harrison, Hewett, Javaheripi, Kauffmann et~al.}]{abdin2024phi}
Marah Abdin, Jyoti Aneja, Harkirat Behl, S{\'e}bastien Bubeck, Ronen Eldan, Suriya Gunasekar, Michael Harrison, Russell~J Hewett, Mojan Javaheripi, Piero Kauffmann, and 1 others. 2024.
\newblock Phi-4 technical report.
\newblock \emph{arXiv preprint arXiv:2412.08905}.

\bibitem[{Aguiar et~al.(2024)Aguiar, Zweigenbaum, and Naderi}]{Aguiar2024}
M.~Aguiar, P.~Zweigenbaum, and N.~Naderi. 2024.
\newblock Seme at semeval-2024 task 2: Comparing masked and generative language models on natural language inference for clinical trials.
\newblock \emph{arXiv preprint arXiv:2404.03977}.

\bibitem[{Alghatani et~al.(2021)Alghatani, Ammar, Rezgui, Shaban-Nejad et~al.}]{alghatani2021predicting}
Khalid Alghatani, Nariman Ammar, Abdelmounaam Rezgui, Arash Shaban-Nejad, and 1 others. 2021.
\newblock Predicting intensive care unit length of stay and mortality using patient vital signs: machine learning model development and validation.
\newblock \emph{JMIR medical informatics}, 9(5):e21347.

\bibitem[{Altinok(2024)}]{Altinok2024}
D.~Altinok. 2024.
\newblock D-nlp at semeval-2024 task 2: Evaluating clinical inference capabilities of large language models.
\newblock \emph{arXiv preprint arXiv:2405.04170}.

\bibitem[{{American Heart Association}(2024)}]{aha_american_2024}
{American Heart Association}. 2024.
\newblock Understanding blood pressure readings and heart rate guidelines.
\newblock \url{https://www.heart.org/en/health-topics/high-blood-pressure/understanding-blood-pressure-readings}.

\bibitem[{Ankit~Pal(2024)}]{OpenBioLLMs}
Malaikannan~Sankarasubbu Ankit~Pal. 2024.
\newblock Openbiollms: Advancing open-source large language models for healthcare and life sciences.
\newblock \url{https://huggingface.co/aaditya/OpenBioLLM-Llama3-70B}.

\bibitem[{Bakhshandeh(2023)}]{bakhshandeh2023benchmarking}
Sadra Bakhshandeh. 2023.
\newblock Benchmarking medical large language models.
\newblock \emph{Nature Reviews Bioengineering}, 1(8):543--543.

\bibitem[{Beizer and Wiley(1996)}]{beizer1996black}
Boris Beizer and J~Wiley. 1996.
\newblock Black box testing: Techniques for functional testing of software and systems.
\newblock \emph{iEEE Software}, 13(5):98.

\bibitem[{Candel et~al.(2022)Candel, Duijzer, Gaakeer, Ter~Avest, Sir, Lameijer, Hessels, Reijnen, van Zwet, de~Jonge et~al.}]{candel2022association}
Bart~GJ Candel, Ren{\'e}e Duijzer, Menno~I Gaakeer, Ewoud Ter~Avest, {\"O}zcan Sir, Heleen Lameijer, Roger Hessels, Resi Reijnen, Erik~W van Zwet, Evert de~Jonge, and 1 others. 2022.
\newblock The association between vital signs and clinical outcomes in emergency department patients of different age categories.
\newblock \emph{Emergency Medicine Journal}, 39(12):903--911.

\bibitem[{Celi et~al.(2022)Celi, Cellini, Charpignon, Dee, Dernoncourt, Eber, Mitchell, Moukheiber, Schirmer, Situ et~al.}]{celi2022sources}
Leo~Anthony Celi, Jacqueline Cellini, Marie-Laure Charpignon, Edward~Christopher Dee, Franck Dernoncourt, Rene Eber, William~Greig Mitchell, Lama Moukheiber, Julian Schirmer, Julia Situ, and 1 others. 2022.
\newblock Sources of bias in artificial intelligence that perpetuate healthcare disparities—a global review.
\newblock \emph{PLOS Digital Health}, 1(3):e0000022.

\bibitem[{Downey et~al.(2017)Downey, Tahir, Randell, Brown, and Jayne}]{downey2017strengths}
Candice~L Downey, W~Tahir, R~Randell, JM~Brown, and DG~Jayne. 2017.
\newblock Strengths and limitations of early warning scores: a systematic review and narrative synthesis.
\newblock \emph{International journal of nursing studies}, 76:106--119.

\bibitem[{Grattafiori et~al.(2024)Grattafiori, Dubey, Jauhri, Pandey, Kadian, Al-Dahle, Letman, Mathur, Schelten, Vaughan et~al.}]{grattafiori2024llama}
Aaron Grattafiori, Abhimanyu Dubey, Abhinav Jauhri, Abhinav Pandey, Abhishek Kadian, Ahmad Al-Dahle, Aiesha Letman, Akhil Mathur, Alan Schelten, Alex Vaughan, and 1 others. 2024.
\newblock The llama 3 herd of models.
\newblock \emph{arXiv preprint arXiv:2407.21783}.

\bibitem[{Gu et~al.(2021)Gu, Tinn, Cheng, Lucas, Usuyama, Liu, Naumann, Gao, and Poon}]{gu2021domain}
Yu~Gu, Robert Tinn, Hao Cheng, Michael Lucas, Naoto Usuyama, Xiaodong Liu, Tristan Naumann, Jianfeng Gao, and Hoifung Poon. 2021.
\newblock Domain-specific language model pretraining for biomedical natural language processing.
\newblock \emph{ACM Transactions on Computing for Healthcare (HEALTH)}, 3(1):1--23.

\bibitem[{Guo et~al.(2025)Guo, Yang, Zhang, Song, Zhang, Xu, Zhu, Ma, Wang, Bi et~al.}]{guo2025deepseek}
Daya Guo, Dejian Yang, Haowei Zhang, Junxiao Song, Ruoyu Zhang, Runxin Xu, Qihao Zhu, Shirong Ma, Peiyi Wang, Xiao Bi, and 1 others. 2025.
\newblock Deepseek-r1: Incentivizing reasoning capability in llms via reinforcement learning.
\newblock \emph{arXiv preprint arXiv:2501.12948}.

\bibitem[{Hempel et~al.(2023)Hempel, Sadeghi, and Kirsten}]{hempel2023prediction}
Lars Hempel, Sina Sadeghi, and Toralf Kirsten. 2023.
\newblock Prediction of intensive care unit length of stay in the mimic-iv dataset.
\newblock \emph{Applied Sciences}, 13(12):6930.

\bibitem[{Herasevich et~al.(2022)Herasevich, Lipatov, Pinevich, Lindroth, Tekin, Herasevich, Pickering, and Barwise}]{herasevich2022impact}
Svetlana Herasevich, Kirill Lipatov, Yuliya Pinevich, Heidi Lindroth, Aysun Tekin, Vitaly Herasevich, Brian~W Pickering, and Amelia~K Barwise. 2022.
\newblock The impact of health information technology for early detection of patient deterioration on mortality and length of stay in the hospital acute care setting: systematic review and meta-analysis.
\newblock \emph{Critical care medicine}, 50(8):1198--1209.

\bibitem[{{Infectious Diseases Society of America}(2003)}]{idsa_fever_2003}
{Infectious Diseases Society of America}. 2003.
\newblock Fever and neutropenia clinical practice guidelines.
\newblock \url{https://www.idsociety.org/practice-guideline/febrile-neutropenia/}.

\bibitem[{Johnson et~al.(2023{\natexlab{a}})Johnson, Pollard, Horng, Celi, and Mark}]{Johnson2023-gt}
Alistair Johnson, Tom Pollard, Steven Horng, Leo~Anthony Celi, and Roger Mark. 2023{\natexlab{a}}.
\newblock {MIMIC-IV-Note}: Deidentified free-text clinical notes.

\bibitem[{Johnson et~al.(2023{\natexlab{b}})Johnson, Bulgarelli, Shen, Gayles, Shammout, Horng, Pollard, Hao, Moody, Gow et~al.}]{johnson2023mimic}
Alistair~EW Johnson, Lucas Bulgarelli, Lu~Shen, Alvin Gayles, Ayad Shammout, Steven Horng, Tom~J Pollard, Sicheng Hao, Benjamin Moody, Brian Gow, and 1 others. 2023{\natexlab{b}}.
\newblock Mimic-iv, a freely accessible electronic health record dataset.
\newblock \emph{Scientific data}, 10(1):1.

\bibitem[{Jullien et~al.(2024)Jullien, Valentino, and Freitas}]{Jullien2024}
M.~Jullien, M.~Valentino, and A.~Freitas. 2024.
\newblock Semeval-2024 task 2: Safe biomedical natural language inference for clinical trials.
\newblock \emph{arXiv preprint arXiv:2404.04963}.

\bibitem[{Kesen et~al.(2024)Kesen, Pedrotti, Dogan, Cafagna, Acikgoz, Parcalabescu, Calixto, Frank, Gatt, Erdem, and Erdem}]{kesen2023vilma}
Ilker Kesen, Andrea Pedrotti, Mustafa Dogan, Michele Cafagna, Emre~Can Acikgoz, Letitia Parcalabescu, Iacer Calixto, Anette Frank, Albert Gatt, Aykut Erdem, and Erkut Erdem. 2024.
\newblock Vilma: A zero-shot benchmark for linguistic and temporal grounding in video-language models.
\newblock In \emph{International Conference on Learning Representations (ICLR)}.

\bibitem[{Kougia et~al.(2024)Kougia, Sedova, Stephan, Zaporojets, and Roth}]{Kougia2024}
V.~Kougia, A.~Sedova, A.~Stephan, K.~Zaporojets, and B.~Roth. 2024.
\newblock Analysing zero-shot temporal relation extraction on clinical notes using temporal consistency.
\newblock \emph{arXiv preprint arXiv:2406.11486}.

\bibitem[{Lee et~al.(2025)Lee, Shang, Baik, Duong-Tran, Yang, Li, and Shen}]{lee2025investigating}
Joseph Lee, Tianqi Shang, Jae~Young Baik, Duy Duong-Tran, Shu Yang, Lingyao Li, and Li~Shen. 2025.
\newblock Investigating llms in clinical triage: Promising capabilities, persistent intersectional biases.
\newblock \emph{arXiv preprint arXiv:2504.16273}.

\bibitem[{Macias-Konstantopoulos et~al.(2023)Macias-Konstantopoulos, Collins, Diaz, Duber, Edwards, Hsu, Ranney, Riviello, Wettstein, and Sachs}]{Macias-Konstantopoulos2023-xi}
Wendy~L Macias-Konstantopoulos, Kimberly~A Collins, Rosemarie Diaz, Herbert~C Duber, Courtney~D Edwards, Antony~P Hsu, Megan~L Ranney, Ralph~J Riviello, Zachary~S Wettstein, and Carolyn~J Sachs. 2023.
\newblock Race, healthcare, and health disparities: A critical review and recommendations for advancing health equity.
\newblock \emph{West J Emerg Med}, 24(5):906--918.

\bibitem[{MacPhail et~al.(2024)MacPhail, Harbecke, Raithel, and Möller}]{MacPhail2024}
D.~MacPhail, D.~Harbecke, L.~Raithel, and S.~Möller. 2024.
\newblock Evaluating the robustness of adverse drug event classification models using templates.
\newblock \emph{arXiv preprint arXiv:2407.02432}.

\bibitem[{McDuff et~al.(2023)McDuff, Schaekermann, Tu, Palepu, Wang, Garrison, Singhal, Sharma, Azizi, Kulkarni et~al.}]{mcduff2023towards}
Daniel McDuff, Mike Schaekermann, Tao Tu, Anil Palepu, Amy Wang, Jake Garrison, Karan Singhal, Yash Sharma, Shekoofeh Azizi, Kavita Kulkarni, and 1 others. 2023.
\newblock Towards accurate differential diagnosis with large language models.
\newblock \emph{arXiv preprint arXiv:2312.00164}.

\bibitem[{OpenAI(2025)}]{openai2025gptoss120bgptoss20bmodel}
OpenAI. 2025.
\newblock \href {https://arxiv.org/abs/2508.10925} {gpt-oss-120b \& gpt-oss-20b model card}.
\newblock \emph{Preprint}, arXiv:2508.10925.

\bibitem[{{OpenAI(gpt-4.1-mini)}(2025)}]{openai_gpt41mini}
{OpenAI(gpt-4.1-mini)}. 2025.
\newblock Gpt-4.1 mini.
\newblock \url{https://platform.openai.com/docs/models/gpt-4.1-mini}.
\newblock Large language model.

\bibitem[{{OpenMeditron}(2025)}]{meditron3phi4}
{OpenMeditron}. 2025.
\newblock Meditron-3-phi4-14b.
\newblock \url{https://huggingface.co/OpenMeditron/Meditron3-Phi4-14B}.
\newblock Accessed: 2025-05-13.

\bibitem[{Parcalabescu et~al.(2022)Parcalabescu, Cafagna, Muradjan, Frank, Calixto, and Gatt}]{parcalabescu-etal-2022-valse}
Letitia Parcalabescu, Michele Cafagna, Lilitta Muradjan, Anette Frank, Iacer Calixto, and Albert Gatt. 2022.
\newblock \href {https://doi.org/10.18653/v1/2022.acl-long.567} {{VALSE}: A task-independent benchmark for vision and language models centered on linguistic phenomena}.
\newblock In \emph{Proceedings of the 60th Annual Meeting of the Association for Computational Linguistics (Volume 1: Long Papers)}, pages 8253--8280, Dublin, Ireland. Association for Computational Linguistics.

\bibitem[{Ribeiro et~al.(2020)Ribeiro, Wu, Guestrin, and Singh}]{ribeiro2020beyond}
Marco~Tulio Ribeiro, Tongshuang Wu, Carlos Guestrin, and Sameer Singh. 2020.
\newblock Beyond accuracy: Behavioral testing of nlp models with checklist.
\newblock \emph{arXiv preprint arXiv:2005.04118}.

\bibitem[{R{\"o}hr et~al.(2024)R{\"o}hr, Figueroa, Papaioannou, Fallon, Bressem, Nejdl, and L{\"o}ser}]{rohr2024revisiting}
Tom R{\"o}hr, Alexei Figueroa, Jens-Michalis Papaioannou, Conor Fallon, Keno Bressem, Wolfgang Nejdl, and Alexander L{\"o}ser. 2024.
\newblock Revisiting clinical outcome prediction for mimic-iv.
\newblock In \emph{Proceedings of the 6th Clinical Natural Language Processing Workshop}, pages 208--217.

\bibitem[{{Royal College of Physicians}(2017)}]{news2_2017}
{Royal College of Physicians}. 2017.
\newblock National early warning score (news) 2.
\newblock \url{https://www.rcplondon.ac.uk/projects/outputs/national-early-warning-score-news-2}.

\bibitem[{Singhal et~al.(2025)Singhal, Tu, Gottweis, Sayres, Wulczyn, Amin, Hou, Clark, Pfohl, Cole-Lewis et~al.}]{singhal2025toward}
Karan Singhal, Tao Tu, Juraj Gottweis, Rory Sayres, Ellery Wulczyn, Mohamed Amin, Le~Hou, Kevin Clark, Stephen~R Pfohl, Heather Cole-Lewis, and 1 others. 2025.
\newblock Toward expert-level medical question answering with large language models.
\newblock \emph{Nature Medicine}, pages 1--8.

\bibitem[{{Society of Critical Care Medicine}(2015)}]{sccm_temp_2015}
{Society of Critical Care Medicine}. 2015.
\newblock Targeted temperature management after cardiac arrest.
\newblock \url{https://www.sccm.org/blog/concise-critical-appraisal-temperature-management-after-cardiac-arrest}.

\bibitem[{{Society of Critical Care Medicine}(2021)}]{sccm_sepsis_2021}
{Society of Critical Care Medicine}. 2021.
\newblock Surviving sepsis campaign: International guidelines for the management of sepsis and septic shock 2021.
\newblock \url{https://www.sccm.org/SurvivingSepsisCampaign/Guidelines/Adult-Patients}.

\bibitem[{Team(2024)}]{qwen2.5}
Qwen Team. 2024.
\newblock \href {https://qwenlm.github.io/blog/qwen2.5/} {Qwen2.5: A party of foundation models}.

\bibitem[{Van~Aken et~al.(2021{\natexlab{a}})Van~Aken, Herrmann, and Löser}]{vanAken2021}
B.~Van~Aken, S.~Herrmann, and A.~Löser. 2021{\natexlab{a}}.
\newblock What do you see in this patient? behavioral testing of clinical nlp models.
\newblock \emph{arXiv preprint arXiv:2111.15512}.

\bibitem[{Van~Aken et~al.(2021{\natexlab{b}})Van~Aken, Papaioannou, Mayrdorfer, Budde, Gers, and Loeser}]{van2021clinical}
Betty Van~Aken, Jens-Michalis Papaioannou, Manuel Mayrdorfer, Klemens Budde, Felix~A Gers, and Alexander Loeser. 2021{\natexlab{b}}.
\newblock Clinical outcome prediction from admission notes using self-supervised knowledge integration.
\newblock \emph{arXiv preprint arXiv:2102.04110}.

\bibitem[{Van~Veen et~al.(2024)Van~Veen, Van~Uden, Blankemeier, Delbrouck, Aali, Bluethgen, Pareek, Polacin, Reis, Seehofnerov{\'a} et~al.}]{van2024adapted}
Dave Van~Veen, Cara Van~Uden, Louis Blankemeier, Jean-Benoit Delbrouck, Asad Aali, Christian Bluethgen, Anuj Pareek, Malgorzata Polacin, Eduardo~Pontes Reis, Anna Seehofnerov{\'a}, and 1 others. 2024.
\newblock Adapted large language models can outperform medical experts in clinical text summarization.
\newblock \emph{Nature medicine}, 30(4):1134--1142.

\bibitem[{{World Health Organization}(2016)}]{who_hypoxemia_2016}
{World Health Organization}. 2016.
\newblock Oxygen therapy for children: A manual for health workers.
\newblock \url{https://www.who.int/publications/i/item/9789241549554}.

\bibitem[{Xu et~al.(2023)Xu, Lu, Yang, Liang, Peng, Pang, Ding, Shi, Yang, Song et~al.}]{xu2023medgpteval}
Jie Xu, Lu~Lu, Sen Yang, Bilin Liang, Xinwei Peng, Jiali Pang, Jinru Ding, Xiaoming Shi, Lingrui Yang, Huan Song, and 1 others. 2023.
\newblock Medgpteval: A dataset and benchmark to evaluate responses of large language models in medicine.
\newblock \emph{arXiv preprint arXiv:2305.07340}.

\bibitem[{Yao et~al.(2024)Yao, Zhang, Tang, Bian, Zhao, Yang, Wang, Zhou, Jang, Ouyang et~al.}]{yao2024medqa}
Zonghai Yao, Zihao Zhang, Chaolong Tang, Xingyu Bian, Youxia Zhao, Zhichao Yang, Junda Wang, Huixue Zhou, Won~Seok Jang, Feiyun Ouyang, and 1 others. 2024.
\newblock Medqa-cs: Benchmarking large language models clinical skills using an ai-sce framework.
\newblock \emph{arXiv preprint arXiv:2410.01553}.

\bibitem[{Zack et~al.(2024)Zack, Lehman, Suzgun, Rodriguez, Celi, Gichoya, and Alsentzer}]{Zack2024}
T.~Zack, E.~Lehman, M.~Suzgun, J.~A. Rodriguez, L.~A. Celi, J.~Gichoya, and E.~Alsentzer. 2024.
\newblock Assessing the potential of gpt-4 to perpetuate racial and gender biases in health care: a model evaluation study.
\newblock \emph{The Lancet Digital Health}, 6(1):e12--e22.

\bibitem[{Zhao et~al.(2024)Zhao, Wang, Liu, Suhuang, Wu, and Zheng}]{Zhao2024}
Y.~Zhao, H.~Wang, Y.~Liu, W.~Suhuang, X.~Wu, and Y.~Zheng. 2024.
\newblock Can llms replace clinical doctors? exploring bias in disease diagnosis by large language models.
\newblock In \emph{Findings of the Association for Computational Linguistics: EMNLP 2024}, pages 13914--13935.

\end{thebibliography}

% uncomment to add appendix
\appendix
\section{Appendix}
\label{sec:appendix}

\subsection{Raw Clinical Note Examples}
\label{sec:demo-notes-raw}

\parahead{Correct Discharge Note (Raw)}

\noindent\fbox{\begin{minipage}{\columnwidth}
\textsf{\small{
\textbf{PRESENT ILLNESS:} The patient is a \_\_\_ year-old female with a history of NSCLC (stage IV) who presents with shortness of breath. (\dots)\\
\textbf{MEDICAL HISTORY:} CAD s/p MI \_\_\_, s/p CABG \_\_\_, Hypertension, Dyslipidemia, CVA: small left posterior frontal infarct in \_\_\_, Macular Degeneration, NSCLC- stage IV. (\dots)\\
\textbf{MEDICATION ON ADMISSION:} amlodipine 5 mg, atorvastatin [Lipitor] 80 mg, calcitriol 0.25 mcg, clopidogrel [Plavix] 75 mg, folic acid 1 mg, furosemide 40 mg. (\dots)\\
\textbf{ALLERGIES:} Codeine\\
\textbf{PHYSICAL EXAM:} On Admission: Vitals: T: 96.9, BP: 118/51, \textcolor{blue}{\textbf{HR: 94}} , RR: 18, O2Sat: 94\% on 5L with face tent.\\
\textbf{FAMILY HISTORY:} Father died of CAD; mother had stomach cancer and osteosarcoma.\\
\textbf{SOCIAL HISTORY:} \_\_\_.}}
\end{minipage}}\\

\parahead{Counterfactual Discharge Note (Raw)}
Identical to the correct note, except for the heart rate in \textbf{\textsf{\small{PHYSICAL EXAM}}}, which is replaced by \textcolor{red}{\textsf{\small{\textbf{HR: 120}}}} instead of \textcolor{blue}{\textsf{\small{\textbf{HR: 94}}}}. We do not reproduce the entire note to avoid clutter.

\subsection{Template-based clinical note examples}
\label{sec:demo-notes-templatebased}
\parahead{Correct Discharge Note (Template)}
The original discharge note example shown in template-based format: 

\noindent\fbox{\begin{minipage}{\columnwidth}
\textsf{\small{
Age: 67\\
Gender: F\\
Ethnicity: White\\
Vitals:\\
\quad \textcolor{blue}{\textbf{Heart Rate: 94}}\\
\quad Blood Pressure: 118/51\\
\quad Respiration Rate: 18\\
\quad Temperature: 96.9\\
\quad Oxygen Saturation: 94\%
}}
\end{minipage}}\\

\parahead{Counterfactual Discharge Note (Template)}
%\noindent\textbf{Counterfactual on heart rate (structured):}

\noindent\fbox{\begin{minipage}{\columnwidth}
\textsf{\small{
Age: 67\\
Gender: F\\
Ethnicity: White\\
Vitals:\\
\quad \textcolor{red}{\textbf{Heart Rate: 120}}\\
\quad Blood Pressure: 118/51\\
\quad Respiration Rate: 18\\
\quad Temperature: 96.9\\
\quad Oxygen Saturation: 94\%
}}
\end{minipage}}\\

%\subsection{Fine-tuning Specification}
%\label{sec:FT-specifications}
%Only admission-time sections were retained to avoid leakage of future information. Fine-tuning was conducted with the LLaMA Factory framework using LoRA adapters, trained over 3 epochs (learning rate: 5e-5, batch size: 2, gradient accumulation: 4). Evaluation was performed every 10 steps using a held-out validation set.

\subsection{Extraction of vital signs Specification}
\label{sec:extraction-specifications}

To extract vital signs from unstructured clinical notes, we use a few-shot prompt applied to the \texttt{PHYSICAL EXAM} section of each note. The model is instructed to extract the vital signs when present.

\textbf{Prompt template:}

{\small
\setlength{\parskip}{0pt}

You are a clinical information extraction assistant. Your task is to extract the vitals from the PHYSICAL EXAM section of a clinical note.

\begin{itemize}[noitemsep,topsep=0pt,parsep=0pt,partopsep=0pt,leftmargin=*]
    \item Temperature
    \item Heart Rate (or Pulse)
    \item Blood Pressure
    \item Respiration Rate
    \item Oxygen Saturation
\end{itemize}

The vitals text may present these values in various formats.

\textbf{Example Input 1:}
\begin{Verbatim}[fontsize=\small,breaklines=true,breakanywhere=true]
{"subject_id":"12345","hamd_id":"abcde","section":"PHYSICAL EXAM","content":"Vitals: T 97.7, HR 110, BP 99/62, RR 25, O2 99%"}
\end{Verbatim}

\textbf{Example Output 1:}
\begin{Verbatim}[fontsize=\small,breaklines=true,breakanywhere=true]
{"subject_id":"12345","hamd_id":"abcde","vitals":{"temperature":"97.7","heart_rate":"110","blood_pressure":"99/62","respiration_rate":"25","oxygen_saturation":"99%"}}
\end{Verbatim}

\textbf{Example Input 2:}
\begin{Verbatim}[fontsize=\small,breaklines=true,breakanywhere=true]
{"subject_id":"12347","hamd_id":"abcde","section":"PHYSICAL EXAM","content":"T 98.2, P 117, O2 97%"}
\end{Verbatim}

\textbf{Example Output 2:}
\begin{Verbatim}[fontsize=\small,breaklines=true,breakanywhere=true]
{"subject_id":"12347","hamd_id":"abcde","vitals":{"temperature":"98.2","heart_rate":"117","blood_pressure":"","respiration_rate":"","oxygen_saturation":"97%"}}
\end{Verbatim}

\noindent IMPORTANT: Return ONLY the JSON object and nothing else.
}

\subsection{Clinical guideline-based severity scales}
\label{sec:clinical_guidelines}
The goal of the severity scale is to create a standardized categorical representation of vital signs that enables aggregation, comparison, and interpretation across variables and experiments.
Vital signs differ substantially in their numerical ranges, units, and clinical interpretation. To support a unified counterfactual evaluation framework, we map each continuous vital sign to a shared set of severity categories (\texttt{very low}, \texttt{low}, \texttt{normal}, \texttt{high}, \texttt{very high}). This standardization allows results to be aggregated across vital signs while preserving clinically meaningful magnitude and directionality of change. 
We manually derive severity ranges for each vital sign from established clinical guidelines~\cite{idsa_fever_2003,news2_2017,sccm_temp_2015,who_hypoxemia_2016,sccm_sepsis_2021,aha_american_2024}.
These variables are widely used in early warning, with ranges chosen to reflect clinically interpretable transitions between physiological states rather than precise diagnostic cutoffs. The resulting categories therefore represent suitable severity levels for the understanding of behavioral analyses.

%\subsection{Evaluation Metrics Aggregation Details}
%\label{sec:ev-metrics-details}
%\textbf{Jensen–Shannon divergence (JSD)} and the \textbf{Expected LOS shift ($\Delta \mathbb{E}_\text{LOS}$)} are aggregated per admission (by \texttt{hadm\_id}), then averaged across counterfactuals for each variable to ensure comparability. We further group results by model family (e.g., fine-tuned vs. zero-shot) and perturbation severity.

\subsection{Full Population Statistics}
Table~\ref{tab:population_stats} presents cohort statistics, including demographics, vitals, and missing data percentages.

\begin{table}[t!]
\centering
\resizebox{0.48\textwidth}{!}{
\begin{tabular}{lll}
\toprule
\textbf{Variable} & \textbf{Value} & \textbf{Missing (\%)} \\
\midrule
Sex (M / F)              & 55.2\% / 44.8\%             & 0\% \\
\midrule
Race                    & White: 69.1\%               & 0\% \\
                        & Other/Unknown: 13.2\%       &     \\
                        & Black: 10.2\%               &     \\
                        & Asian/Pacific: 4.0\%        &     \\
                        & Hispanic/Latino: 3.5\%      &     \\
\midrule
Age (years)             & 63.64 ± 16.85               & 0\% \\
\midrule
Temperature (°F)        & 97.10 ± 7.69                & 27\% \\
Heart rate (bpm)        & 83.86 ± 20.47               & 5.7\% \\
Respiration rate (bpm)  & 18.93 ± 5.38                & 12\% \\
Oxygen saturation (\%)  & 96.99 ± 3.49                & 5.1\% \\
Systolic BP (mmHg)      & 128.90 ± 24.15              & 4.3\% \\
Diastolic BP (mmHg)     & 71.03 ± 15.46               & 4.3\% \\
\bottomrule
\end{tabular}}
\caption{Cohort-level statistics used for counterfactual testing, including demographics, vitals, and percentage of missing values.}
\label{tab:population_stats}
\end{table}

%\subsection{LOS class representation}
%\label{app:LOS_class_representation}
\subsection{Zero-shot prompts}
\label{app:zero-shot-prompts}
We evaluate models in a zero-shot setting on two downstream tasks: ICU length-of-stay (LoS) prediction and mortality prediction. 

\paragraph{ICU length-of-stay (LoS) prediction.}
Models are asked to predict the LoS class based on the patient's admission note.

%\textbf{Prompt template (LoS):}
\vspace{0.3cm}
{\small
\setlength{\parskip}{0pt}

You are an expert in ICU length-of-stay prediction.
Based only on the patient's admission note, predict in which ICU length-of-stay bucket the patient will fall.

We divide ICU length of stay (for stays $\geq 24$ hours) into 4 groups:
\begin{itemize}[noitemsep,topsep=0pt,parsep=0pt,partopsep=0pt,leftmargin=*]
    \item \texttt{[[1]]} Very short stay (24 to 38 hours)
    \item \texttt{[[2]]} Short stay (39 to 59 hours)
    \item \texttt{[[3]]} Moderate stay (60 to 112 hours)
    \item \texttt{[[4]]} Long stay (113 to 657 hours)
\end{itemize}

\noindent Return only the bucket in double brackets: \texttt{[[1]]}, \texttt{[[2]]}, \texttt{[[3]]}, or \texttt{[[4]]}.
}

\paragraph{In-hospital mortality prediction.}
Models are asked to predict whether the patient will die during the hospital admission.

%\textbf{Prompt template (mortality):}
\vspace{0.3cm}
{\small
\setlength{\parskip}{0pt}

You are an expert in ICU mortality prediction.
Based only on the patient's structured admission note, predict whether the patient will die before ICU discharge.

Return only the target value in double brackets:
\begin{itemize}[noitemsep,topsep=0pt,parsep=0pt,partopsep=0pt,leftmargin=*]
    \item \texttt{[[0]]} for survival
    \item \texttt{[[1]]} for death
\end{itemize}
}

%We provide additional results about LOS prediction in Table~\ref{tab:los_distribution}.

%\begin{table}[t!]
%\centering
%\resizebox{0.20\textwidth}{!}{
%\begin{tabular}{ll}
%\toprule
%\textbf{Bucket} & \textbf{\# Cases} \\
%\midrule
%1  & 30674 \\
%2  & 10028 \\
%3  & 3872  \\
%4  & 2177  \\
%\bottomrule
%\end{tabular}}
%\caption{LOS class distribution in the training set. There is an over-representation of shorter stays making it harder for the FT models to predict longer stay buckets.}
%\label{tab:los_distribution}
%\end{table}

\subsection{Results}
\subsubsection{Model Specifications}
\label{app:llm-models}
Table~\ref{tab:llm-models} summarizes the LLMs used, including model type, domain, and background info on training.

\begin{table*}[t!]
\centering
\small
\begin{tabularx}{\textwidth}{p{3.0cm}p{2.5cm}rp{2.7cm}X}
\toprule
\textbf{LLM} & \textbf{Base LLM} & \textbf{\#Params.} & \textbf{Domain} & \textbf{Notes} \\
\midrule
OpenBioLLM\newline{\footnotesize \cite{OpenBioLLMs}} 
& Llama-3.3-70B-Instruct & 70B & Biomedical
& Outperforms GPT-4, Gemini, Meditron, and Med-PaLM-2 on biomedical benchmarks. \\
Meditron3-Phi4\newline{\footnotesize \cite{meditron3phi4}} 
& Phi-4 & 14B & Biomedical
& Finetuned version of Phi-4 on medical corpora. \\
Phi 4\newline{\footnotesize \cite{abdin2024phi}} 
& Phi-4 & 14B & General-purpose  
& Trained for efficient language understanding and reasoning. \\
Llama 3.3 Instruct\newline{\footnotesize \cite{grattafiori2024llama}} 
& Llama-3.3-70B-Instruct & 70B & General-purpose 
& Instruction-tuned. Strong performance on general reasoning and text-based tasks. \\
DeepSeek R1 Distill\newline{\footnotesize \cite{guo2025deepseek}} 
& Llama-3.3-70B-Instruct & 70B & General-purpose (reasoning-focused)
& Distilled from DeepSeek-R1 using Llama 3.3-70B-Instruct; optimized for multi-step reasoning.\\
Qwen 2.5 Instruct\newline{\footnotesize \cite{qwen2.5}} 
& Qwen-2.5-72B-Instruct & 72B & General-purpose 
& Strong open-source baseline, with competitive performance relative to other large instruction-tuned LLMs.\\
GPT OSS\newline{\footnotesize \cite{openai2025gptoss120bgptoss20bmodel}} 
& GPT-OSS-120B & 120B & General-purpose 
& Open-weight GPT model. Large-scale alternative to proprietary LLMs, enabling comparison between open and closed models.\\
GPT 4.1 mini\newline{\footnotesize \cite{openai_gpt41mini}} 
& GPT-4.1-mini & N/A & General-purpose 
& Proprietary model serving as a lightweight reference for commercial systems, included to contrast open-weight models.\\
\bottomrule
\end{tabularx}
\caption{List of large language models (LLMs) evaluated in this study, including architecture, domain specialization, and training context.}
\label{tab:llm-models}
\end{table*}

\begin{table}[t!]
\centering
\resizebox{\columnwidth}{!}{%
%\scriptsize
%\setlength{\tabcolsep}{2.5pt}
%\renewcommand{\arraystretch}{1.05}
\begin{tabular}{lrrrrrrrr}
\toprule
& \multicolumn{4}{c}{\textbf{Mort}} & \multicolumn{4}{c}{\textbf{LoS}} \\
\cmidrule(lr){2-5}\cmidrule(lr){6-9}
\textbf{Model} & \textbf{Acc.} & \textbf{F1} & \textbf{Prec.} & \textbf{Rec.}
              & \textbf{Acc.} & \textbf{F1} & \textbf{Prec.} & \textbf{Rec.} \\
\midrule
LR       & 0.65 & 0.21 & 0.13 & 0.66 & --   & --   & --   & --   \\
XGBoost        & 0.74 & 0.22 & 0.14 & 0.52 & 0.29 & 0.29 & 0.30 & 0.29 \\
RF  & 0.93 & 0.00 & 0.00 & 0.00 & 0.28 & 0.28 & 0.28 & 0.28 \\
\bottomrule
\end{tabular}}
\caption{Baseline performance for mortality (Mort) and length-of-stay (LoS). \textbf{LR}: Logistic regression. \textbf{RF}: Random forest. LR is not applicable to LoS (shown as --).}
\label{tab:baselines}
\end{table}

\subsubsection{Additional results}
\label{app:additional-results}
Here we present results of additional experiments and for all models. % overpowering the scale, unabling to analyze the behavior of the rest. 
We first report experiments for standard machine learning models---logistic regression (LR), XGBoost, random forest (RF)---when trained on template-based data in Table~\ref{tab:baselines}.
We observe similar performance between LLMs and these machine learning models on both  tasks.

In Table ~\ref{tab:model_summaries_pr} we report precision and recall per model, task and note modality.

\begin{figure}[!t]
    \centering
    \includegraphics[width=\linewidth]{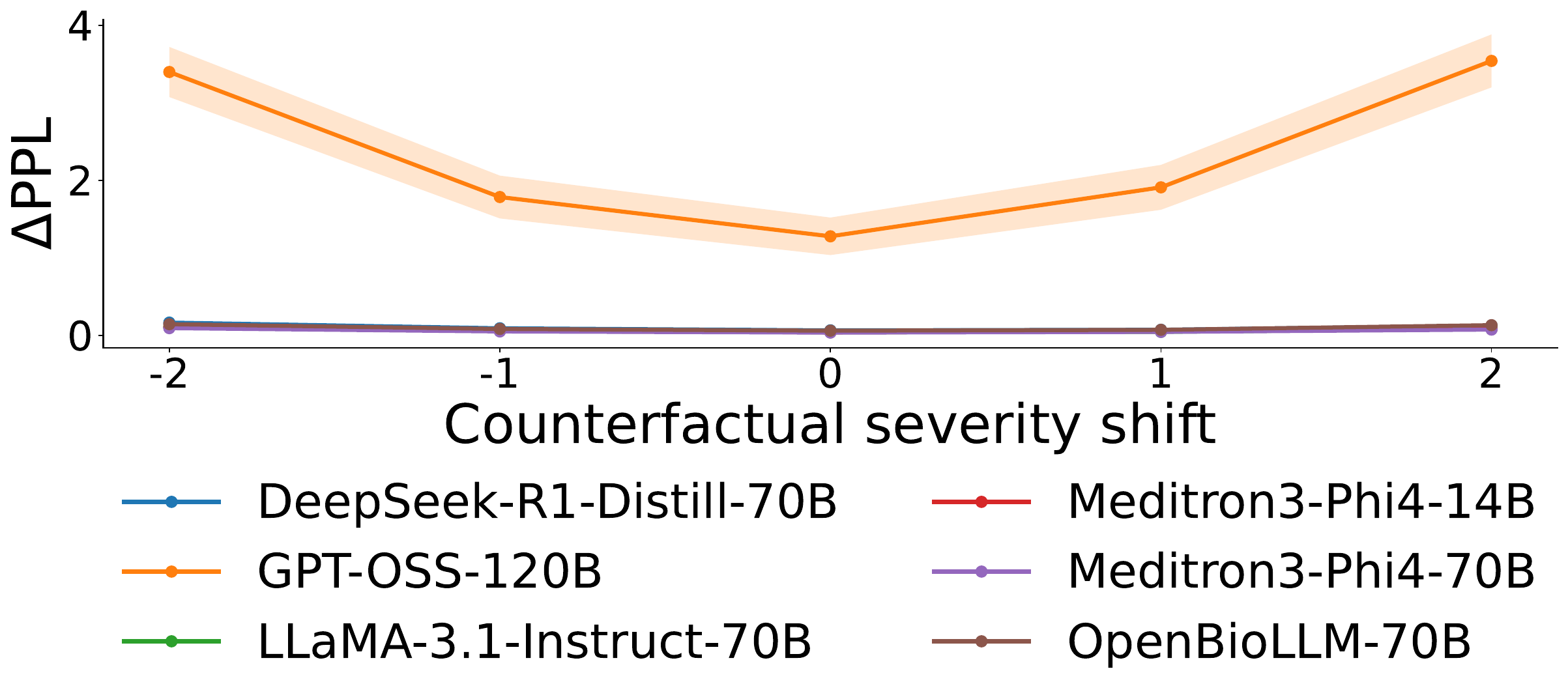}
    \caption{Average per-token $\Delta$PPL across counterfactual severity shifts (raw notes).
    %Positive severity shift: cf is more severe than the original value; negative severity shift: cf is less severe. 
    $\Delta$PPL grows with both increasing and decreasing severity, indicating consistent linguistic sensitivity. \gptoss presents a substantially higher effect ($2.5\pm5.1$). \gptmini is omitted because we could not obtain per-token log probabilities for this model.}
    \label{fig:notask-gpt}
\end{figure}

In Figure~\ref{fig:notask-gpt} we report task-independent experiments with LLMs, including \gptoss. \gptoss shares the same behavior as other LLMs, but with a higher magnitude. 

\begin{figure}[!t]
    \centering
    \includegraphics[width=\linewidth]{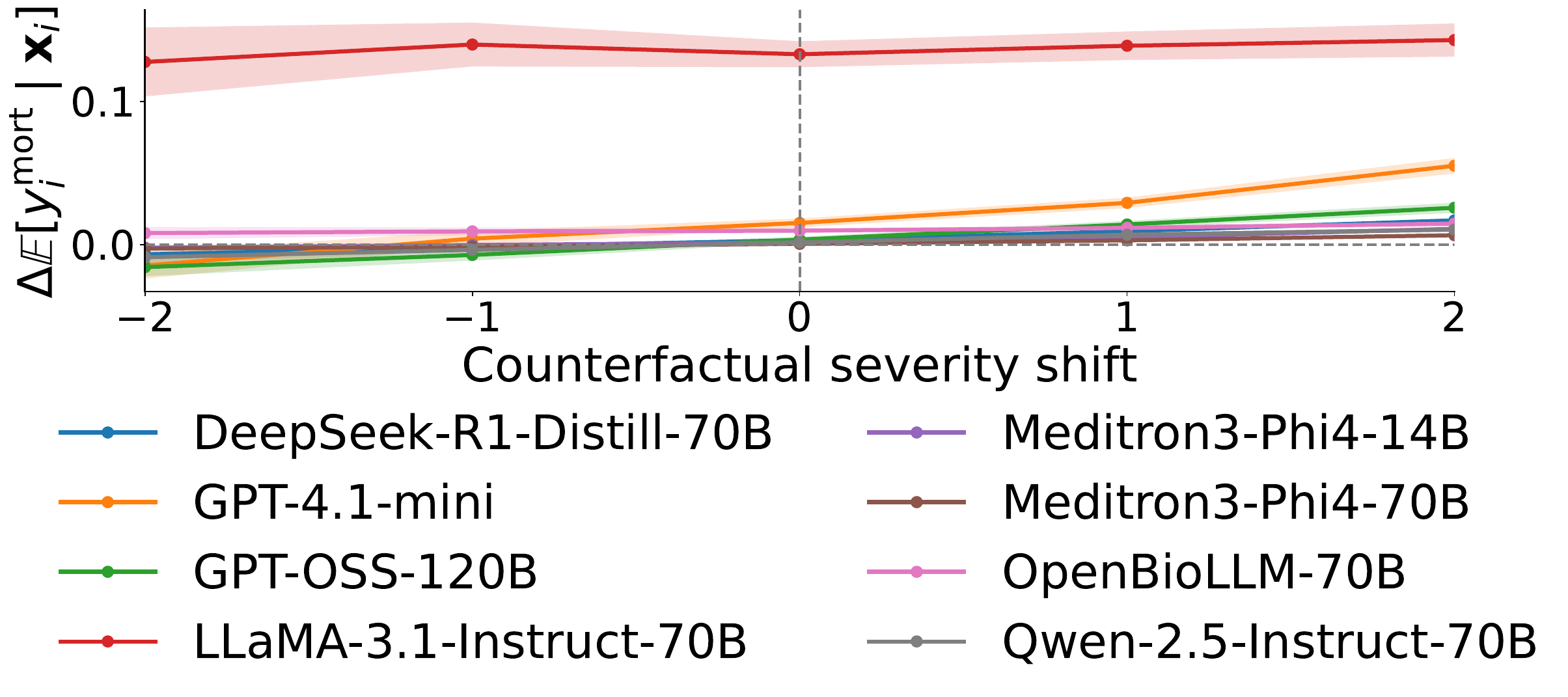}
    \caption{Change of probability of mortality as a function of counterfactual severity shift. Positive severity shifts are expected to increase predicted mortality risk, while negative shifts are expected to decrease it. Most models follow this monotonic trend, indicating clinically aligned responses to vital sign counterfactual perturbations.
    %Llama, due to its substantially larger and different shaped response dominates the scale obscuring the trend of other models.
    }
    \label{fig:dp_mort_llama}
\end{figure}

In Figure~\ref{fig:dp_mort_llama} we report change probability of mortality as a function of coun-
terfactual severity shift, including \llama. This model presented a considerably higher effect than other models and with a different behaviour, overreacting to vital sign perturbations, even whenthe severity stayed the same as the original.
%\begingroup
%\setlength{\abovecaptionskip}{2pt}
%\setlength{\belowcaptionskip}{0pt}
%\setlength{\textfloatsep}{6pt}

\begin{table}[t!]
\centering
{\scriptsize %+
\renewcommand{\arraystretch}{0.80} %+
\setlength{\tabcolsep}{2pt} %b

\begin{tabular}{l c c} 
\hline
\textbf{Model} & \textbf{Prec} & \textbf{Rec} \\
\hline

% ===================== Mortality -- raw =====================
\multicolumn{3}{l}{\textbf{Mortality -- raw notes}} \\
\hline
\rowcolor{reasoncolor}
\deepseek & 0.55 & 0.69 \\
\gptmini & 0.56 & 0.73 \\
\gptoss & \textbf{0.61} & 0.67 \\
\llama & 0.57 & 0.73 \\
\rowcolor{medcolor}
\meditron & 0.54 & 0.64 \\
\rowcolor{medcolor}
\obllm & 0.60 & \textbf{0.74} \\
\phiF & \underline{0.53} & \underline{0.58} \\
\qwen & 0.56 & 0.72 \\
\rowcolor{black!6}
\textbf{Average} & 0.57 & 0.69 \\
\hline

% ===================== Mortality -- template =====================
\multicolumn{3}{l}{\textbf{Mortality -- template-based notes}} \\
\hline
\rowcolor{reasoncolor}
\deepseek & 0.54 & 0.59 \\
\gptmini & 0.55 & 0.59 \\
\gptoss & \underline{0.47} & \underline{0.50} \\
\llama & 0.54 & 0.59 \\
\rowcolor{medcolor}
\meditron & 0.56 & 0.60 \\
\rowcolor{medcolor}
\obllm & \textbf{0.63} & 0.57 \\
\phiF & 0.54 & \textbf{0.63} \\
\qwen & 0.59 & 0.60 \\
\rowcolor{black!6}
\textbf{Average} & 0.55 & 0.58 \\
\hline

% ===================== LoS -- raw =====================
\multicolumn{3}{l}{\textbf{LoS -- raw}} \\
\hline
\rowcolor{reasoncolor}
\deepseek & 0.26 & \underline{0.26} \\
\gptmini & 0.34 & 0.32 \\
\gptoss & 0.29 & 0.27 \\
\llama & 0.30 & \textbf{0.31} \\
\rowcolor{medcolor}
\meditron & \underline{0.21} & 0.28 \\
\rowcolor{medcolor}
\obllm & 0.30 & 0.28 \\
\phiF & 0.22 & 0.28 \\
\qwen & \textbf{0.36} & 0.29 \\
\rowcolor{black!6}
\textbf{Average} & 0.29 & 0.29 \\
\hline

% ===================== LoS -- template =====================
\multicolumn{3}{l}{\textbf{LoS -- template}} \\
\hline
\rowcolor{reasoncolor}
\deepseek & 0.22 & 0.26 \\
\gptmini & 0.25 & 0.26 \\
\gptoss & 0.19 & \textbf{0.26} \\
\llama & \textbf{0.25} & 0.24 \\
\rowcolor{medcolor}
\meditron & 0.17 & 0.24 \\
\rowcolor{medcolor}
\obllm & \underline{0.16} & 0.24 \\
\phiF & 0.18 & \underline{0.23} \\
\qwen & \textbf{0.28} & \textbf{0.28} \\
\rowcolor{black!6}
\textbf{Average} & 0.21 & 0.25 \\
\hline

\end{tabular}}
\caption{Precision and recall for each model variant across note types and tasks. Orange: medical LLM; blue: reasoning-oriented; white: general purpose; gray: averages per block.}
\label{tab:model_summaries_pr}
\end{table}

\end{document}